\PassOptionsToPackage{table}{xcolor}
\documentclass[a4paper,fleqn]{cas-dc}

\usepackage[authoryear]{natbib}

\usepackage{mathtools}
\usepackage{amsmath}
\usepackage{caption}
\usepackage{subcaption}
\usepackage{amssymb}
\usepackage[export]{adjustbox}
\usepackage{multirow}
\usepackage{tikz}
\usepackage[flushleft]{threeparttable}
\usepackage{hyperref} 
\usepackage[outline]{contour}
\usepackage{cases}
\usepackage{pifont}
\usepackage{pgf-pie}
\usepackage{cleveref}
\usepackage{color}
\usepackage{pgfplots}
\usepgfplotslibrary{groupplots}
\usepackage{pgfplotstable}
\usepackage{filecontents}

\usepackage[edges]{forest}
\usetikzlibrary{arrows.meta, 
                shadows, shadows.blur}

\tikzset{every shadow/.style={shadow xshift=5pt,shadow yshift=-5pt}}

\usepackage{calc} 
\tikzset{>=latex} 

\usepackage{tkz-kiviat,numprint}
\usetikzlibrary{shapes.geometric, arrows}
\usetikzlibrary{patterns}

\newcommand{\xmark}{\ding{55}}%
\newcommand{\cmark}{\ding{51}}%

\newcommand\ColorBox[1]{\textcolor{#1}{\rule{2ex}{2ex}}}

\usepackage{xspace}
\makeatletter
\DeclareRobustCommand\onedot{\futurelet\@let@token\@onedot}
\def\@onedot{\ifx\@let@token.\else.\null\fi\xspace}

\def\ie{\emph{i.e}\onedot}

\makeatother

\begin{document}
\let\WriteBookmarks\relax
\def\floatpagepagefraction{1}
\def\textpagefraction{.001}
\shorttitle{Review of datasets and techniques in USVs vision}
\shortauthors{L. Trinh  et~al.}

\title [mode = title]{A comprehensive review of datasets and deep learning techniques for vision in Unmanned Surface Vehicles}                      

\author[1]{Linh Trinh}[type=editor,
                        auid=000,bioid=1,
                        orcid=0000-0002-7679-5511]
\ead{linh.trinh@uantwerpen.be}

\credit{Conceptualization, Data Curation, Investigation, Methodology, Writing - Original Draft}

\author[1]{Siegfried Mercelis}[
    orcid=0000-0001-9355-6566
]
\ead{siegfried.mercelis@uantwerpen.be}

\credit{Supervision, Funding acquisition}

\affiliation[1]{organization={IDLab - Faculty of Applied Engineering},
                addressline={University of Antwerp-imec}, 
                city={Antwerpen},
                postcode={2000}, 
                country={Belgium}}

\author[1]{Ali Anwar}[
    orcid=0000-0002-5523-0634
]
\ead{ali.anwar@uantwerpen.be}

\credit{Supervision, Project administration, Validation, Writing - Review and Editing, Funding acquisition}

\begin{abstract}
Unmanned Surface Vehicles (USVs) have emerged as a major platform in maritime operations, capable of supporting a wide range of applications. USVs can help reduce labor costs, increase safety, save energy, and allow for difficult unmanned tasks in harsh maritime environments. With the rapid development of USVs, many vision tasks such as detection and segmentation become increasingly important. Datasets play an important role in encouraging and improving the research and development of reliable vision algorithms for USVs. In this regard, a large number of recent studies have focused on the release of vision datasets for USVs. Along with the development of datasets, a variety of deep learning techniques have also been studied, with a focus on USVs. However, there is a lack of a systematic review of recent studies in both datasets and vision techniques to provide a comprehensive picture of the current development of vision on USVs, including limitations and trends. In this study, we provide a comprehensive review of both USV datasets and deep learning techniques for vision tasks. Our review was conducted using a large number of vision datasets from USVs. We elaborate several challenges and potential opportunities for research and development in USV vision based on a thorough analysis of current datasets and deep learning techniques.


\end{abstract}

\begin{keywords}
unmanned surface vessels \sep datasets \sep deep learning \sep computer vision
\end{keywords}

\maketitle

\section{Background and Motivations}
\subsection{Background}
\begin{figure}[ht]
    \centering
    \includegraphics[width=0.8\columnwidth]{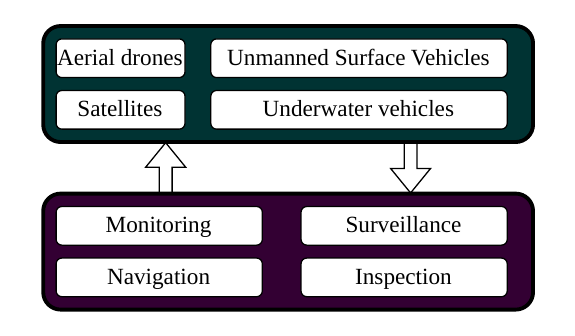}
    \caption{Some platforms and applications in maritime.}
    \label{fig:app_maritime}
\end{figure}

Over 70\% of our planet is comprised of unconfined water environments, which additionally facilitates 80\% of international commerce \citep{review_mt}. Research and development in the maritime sector have been moving very quickly recently. 
Figure \ref{fig:app_maritime} illustrates key maritime platforms and their applications.

\begin{figure}[!b]
    \centering
    \includegraphics[width=0.82\columnwidth]{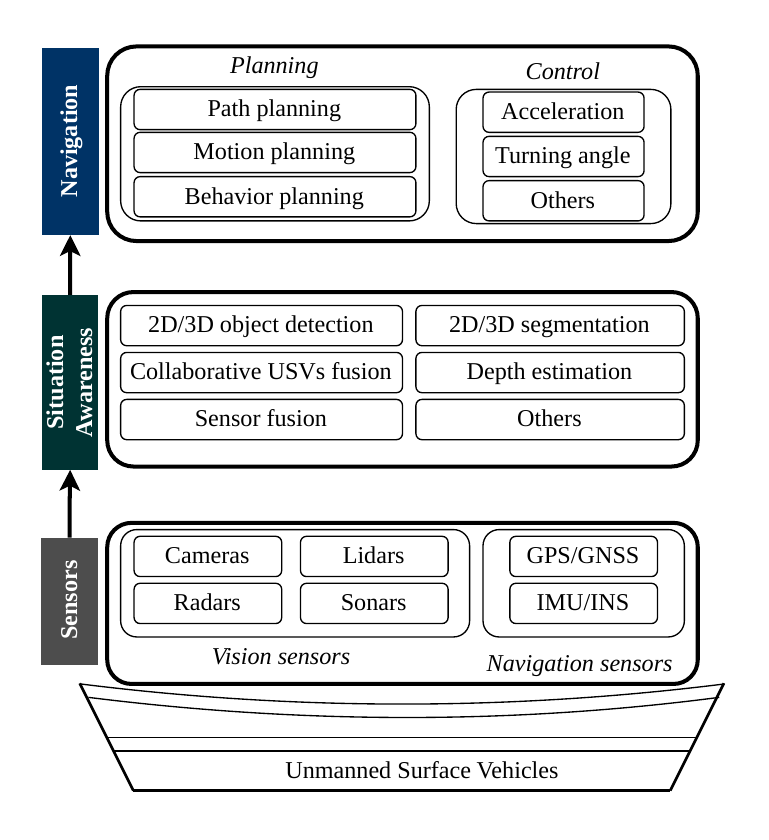}
    \caption{Architecture of a typical USVs.}
    \label{fig:arch_asv}
\end{figure}
There are four main applications in maritime \citep{macvi2023}, which are monitoring, surveillance, inspection and navigation. Several platforms are being developed to support these applications, including aerial drones (UAVs) \citep{survey_vision_data@2023,survey_object_det@2021,survey_small_obj_det_2022,survey_ship_det_2023}, satellite \citep{survey_vision_data@2023,survey_object_det@2021,survey_satellite_2018,survey_ship_det_2023}, underwater vehicles (UWVs) \citep{survey_det@2023}, and unmanned surface vehicles (USVs) \citep{survey_vision_data@2023,survey_object_det@2021}. Unlike UAVs and satellites, which operate above the water without direct interaction with water, and UWVs, which only operate underwater, USVs navigate and perform directly on the water's surface, where they must deal with both atmospheric and marine environmental conditions. This dual-interface capability makes USVs indispensable for comprehensive surface-level data collection, monitoring, surveillance, and intervention \citep{challenge_fov_2016}, roles that neither aerial nor fully submerged systems can perform \citep{maritime_app}. 
As a result, in recent decades, significant efforts have been made to develop USVs that operate on the water's surface. 

 USVs are emerging as autonomous ships and boats. Many research works on USVs were focused on navigation \citep{auv_survey}. Figure \ref{fig:arch_asv} illustrates the common architecture of an autonomous system for the navigation of USVs. Similar to autonomous vehicles (AVs) \citep{av_survey_arch1,av_survey_arch2}, it consists of three major components: sensors, situation awareness, and navigation. Sensors used in vision tasks are classified as either proprioceptive or exteroceptive. Proprioceptive sensors, which detect the state of vessel, include GPS/GNSS, and IMU/INS. These are used to track the platform's position, movement, and odometry. Exteroceptive sensors monitor the environment around vehicles, gathering data on waves, winds, weather, the environment, and external obstacles. This category includes camera, LiDAR, radar, and ultrasonic sensors. 
Depending on the system, various types of cameras have been used, including infrared (IR) cameras, stereo cameras, and panoramic cameras. The autonomous systems like USVs uses these vision sensors to perform a variety of vision tasks, including 2D/3D object detection, segmentation, depth estimation, single/multi-object tracking, and sensor fusions. The primary function of situation awareness in maritime environments is environmental perception, which entails accurately identifying and understanding current surroundings, objects, and events in order to ensure safe navigation and operational decision-making. 
Finally, there is the navigation component, which is primarily responsible for planning and controlling USVs as they navigate the water surface.

\subsection{Motivations and contributions}
\begin{table*}[htbp]
\caption{The comparison of our survey with previous surveys on maritime visions and techniques.(\# datasets: number of collected datasets, last update: last update of dataset or method, OD: object detection, T: object tracking, C: classification, S: segmentation. \xmark: Not mentioned).} 
\label{tab:exist}
\resizebox{\textwidth}{!}{%
\begin{tabular}{|l|l|c|c|c|c|l|}
\hline
\multicolumn{1}{|c|}{\textbf{Study}}    & \multicolumn{1}{c|}{\textbf{Main survey topic}}                      & \textbf{Last update} & \textbf{\# datasets} & \textbf{Tasks}                           & \textbf{Sensors}                & \multicolumn{1}{c|}{\textbf{Target/Source}} \\ \hline
\cite{survey_vision_data@2023}         & Maritime vision datasets                                             & 2021                 & 15                  & 2D OD, T, C                            & RGB camera                             & Satellite, USVs, UAVs                         \\ \hline
\cite{survey_object_det@2021}          & Deep learning-based marine object detection                          & 2021                 & 4                   & 2D OD                                    & RGB camera                             & UAVs, Satellite, USVs                         \\ \hline
\cite{survey_vision@2021}              & Marine vision-based situation awareness using deep learning          & 2021                 & 12                  & 2D OD, 2D S, C, T                      & RGB camera                            & USVs                                         \\ \hline
\cite{survey_det_tracking@2014}        & Object detection and tracking of maritime vessels                    & 2014                 & \xmark                   & 2D OD, T                                 & RGB camera                             & USVs                                         \\ \hline
\cite{survey_small_obj_det_2022}       & Small object detection for maritime surveillance                     & 2022                 & 22                  & 2D OD                                    & RGB camera                            & UAVs, USVs                                    \\ \hline
\cite{survey_ship_classification_2023} & Computer vision architectures for ship classification                & 2023                 & \xmark                   & C                                      & RGB camera                            & N/A                                           \\ \hline
\cite{survey_ship_od_2024}             & Ship maritime object detection based on RGB camera                   & 2023                 & \xmark                   & 2D OD                                    & RGB camera                            & USVs                                         \\ \hline
\cite{survey_satellite_2018}           & Vessel detection and classification from spaceborne optical images   & 2016                 & 119                 & 2D OD, C                               & SAR                             & Satellite                                   \\ \hline
\cite{survey_det_cls_2021}             & Ship detection and classification from optical remote sensing images & 2021                 & 10                  & 2D OD, C                               & RGB camera                             & Map                                         \\ \hline
\cite{survey_ship_det_2023}            & Ship detection with deep learning                                    & 2023                 & 21                  & 2D OD                                    & RGB camera, SAR                         & Satellite, USVs, UAVs                         \\ \hline
\cite{survey_det@2023}                 & Deep learning-based visual detection of maritime organisms           & 2023                 & 25                  & 2D OD                                    & RGB camera                             & UWVs                                         \\ \hline
\textbf{Ours.}                           & Vision datasets and techniques for USVs                               & June 2024             & 38                  & \begin{tabular}[t]{@{}l@{}} 2D/3D OD, 2D/3D S, \\ C, T, Depth, others \end{tabular} & \begin{tabular}[t]{@{}l@{}} RGB camera, Lidar, \\ Radar, Sonar, others \end{tabular} & USVs                                         \\ \hline
\end{tabular}
}
\end{table*}
The application of USVs is heavily dependent on vision capability, especially in busy maritime traffic, near the coast, or in inland waters. These rapid developments in USVs rely heavily on large datasets, which enable USV systems to be robust and reliable in complex environments. To support research and development for maturing USVs, the quality and variety of USVs vision datasets and deep learning techniques have been intensively studied recently. However, there is a lack of a thorough review of research and development of USVs vision to identify current developments, trends, and limitations of both datasets and deep learning techniques. Therefore, in this paper, we carefully review the development of both datasets and deep learning techniques for vision tasks in USVs.

There are several existing surveys on maritime vision datasets and techniques, but not fully focused on USVs. Table \ref{tab:exist} compares our survey with existing works. \cite{survey_det_tracking@2014} early provided a review of vision techniques for maritime vessels. This work focused solely on the object detection and tracking tasks, with no mention of dataset reviews. Furthermore, the discussed techniques are very traditional, such as optical processing for object detection. \cite{survey_satellite_2018} provided a comprehensive review of both datasets and techniques for vessel detection and classification. However, this study focused solely on satellite datasets, specifically synthetic aperture radar (SAR) images. \cite{survey_object_det@2021} focused on deep learning techniques for detecting objects in USVs, UAVs, and satellite platforms using RGB images. Four datasets were used for analysis. \cite{survey_ship_det_2023} discusses deep learning techniques for ship detection and classification using RGB images. For analysis, ten datasets were collected from various map services, including Google Maps and Chinese Maps. \cite{survey_vision@2021} discussed deep learning techniques for various tasks such as object detection, segmentation, classification, and object tracking for USV-based RGB-images. However, the discussed deep learning techniques are limited and predate 2021. They also reviewed 12 datasets related to these tasks. \cite{survey_vision_data@2023} reviewed 15 datasets for maritime vision across three platforms: satellites, USVs, and UAVs. In addition to datasets, there were numerous survey studies on deep learning techniques for object detection, tracking, and classification. \cite{survey_small_obj_det_2022} used 22 small object datasets for the discussion along with several deep learning techniques for small object detection using RGB images for both USVs and UAVs. \cite{survey_ship_classification_2023} discusses the deep learning technique for classification based on RGB-images, but no platform or datasets are mentioned. Recently, several surveys \citep{survey_ship_det_2023,survey_det@2023,survey_ship_od_2024} have primarily focused on the discussion of deep learning techniques for object detection based on RGB images. \cite{survey_vision_data@2023} provided a review of only maritime vision datasets.

Unlike previous surveys, our study focuses on a wide range of vision tasks such as 2D or 3D object detection, segmentation, classification, depth estimation, and many others for USVs. Furthermore, we conduct a systematic review of various sensors for maritime USV vision, including RGB-image (camera, stereo camera, IR-camera), LiDAR, radar, sonar, and others. We provide a comprehensive review of 38 datasets specifically focused on USV vision. To the best of our knowledge, ours is the first study to provide an in-depth review of both USV deep learning techniques as well as datasets. In summary, our contribution can be summarized as below:
\begin{itemize}
    \item We present a comprehensive analysis of a large number of recent datasets for USV vision collected in real-world scenarios using a variety of vision sensors such as cameras, LiDAR, radar, and so on. Overall, 38 public datasets are analyzed for a variety of dataset characteristics. 
    \item We also discuss recent deep learning techniques for USV vision tasks. We carefully categorize the techniques to enable us to perform a thorough analysis.
    \item Finally, based on a thorough review of both datasets and techniques in USV vision, we present several important discussions of the challenges, and potential future perspectives in USVs' vision.
\end{itemize}
We organize the rest of the paper as follows. We present our literature studies on the dataset in Section \ref{sec:dataset}. Then, in Section \ref{sec:technique}, we present the deep review of deep learning techniques on USVs vision. Section \ref{sec:discussion}, discusses our finding, insights and raises the challenges, trends of future works in USVs vision. Finally, we conclude our work in Section \ref{sec:conclude}.


\section{Datasets}\label{sec:dataset}
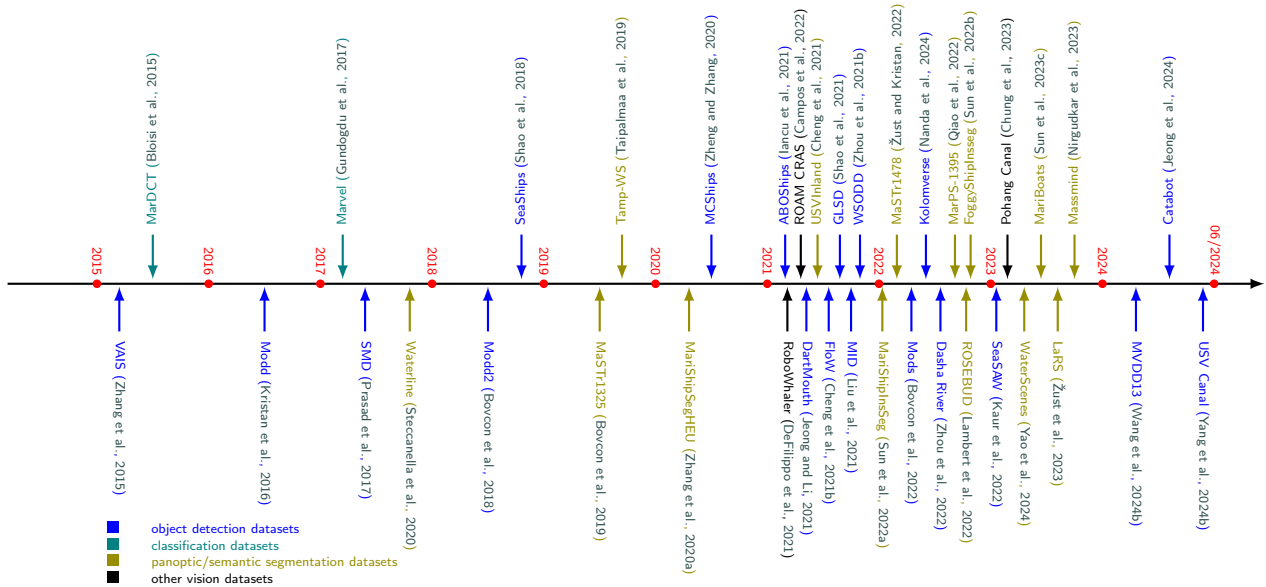
\begin{figure*}[htbp]
    \centering

\begin{tikzpicture}[scale=0.7,font=\tiny] 
 
  \newcount\yearOne; \yearOne=2015
  \newcount\yoffset;
  \def\w{21}       
  \def\n{10}        
  \def\lt{0.20}    
  \def\lf{0.56}    
  \def\lo{0.30}    
  \def\lext{0.07}  
  \def\rext{1.045} 
 
  \def\yearLabel(#1,#2,#3){\node[above,black!60!blue] at ({(#1-\yearOne)*\w/\n/10},{\lt*#2}) {#3};}
  \def\yearArrowLabel(#1,#2,#3,#4,#5){
    \def\xy{{(#1-\yearOne)*\w/\n}}; \pgfmathparse{int(#2*100)};
    \ifnum \pgfmathresult<0 
      \def\yyp{{(\lf*(0.90+#2))}}; \def\yyw{{(\yyp-\lf*#3)}}
      \draw[<-,thick,#5]
        (\xy,\yyp) -- (\xy,\yyw)
        node[below,#5] at (\xy,\yyw) {\strut #4};
    \else 
      \def\yyp{{(\lf*(0.10+#2)}}; \def\yyw{{(\yyp+\lf*#3)}}
      \draw[<-,thick,#5]
        (\xy,\yyp) -- (\xy,\yyw)
        node[above] at (\xy,\yyw) {#4};
    \fi}
  \def\yearArrowLabelRed(#1,#2,#3,#4){
    \def\xy{{(#1-\yearOne)*\w/\n/10}}; \pgfmathparse{int(#2*100)};
      \def\yyp{{(\lt*(0.90+#2))}}; \def\yyw{{(\yyp-\lt*#3)}}
      \fill[red,radius=2pt] (\xy,0) circle;
      \draw[<-,thick,black!25!red,align=center]
        (\xy,\yyp) -- (\xy,\yyw)
        node[below,black!40!red] at (\xy,\yyw) {\strut #4};
     }

 
  \draw[thick,->] (-\w*0.08,0) -- (\w*\rext,0);
 
  \foreach \tick in {0,1,...,\n}{
    \def\x{{\tick*\w/\n}}
    
    \def\year{\the\numexpr \yearOne+\tick*1 \relax}
    \ifnum \tick<\n
        \draw[red,fill=red] (\x,0) circle (.5ex)
                     node[above] {\rotatebox[origin=c]{270}{\year}};
     \else
     \draw[red,fill=red] (\x,0) circle (.5ex)
                     node[above] {\rotatebox[origin=c]{270}{06/2024}};
     \fi

  }
 
 
  \yearArrowLabel(2015.20,-1.0,1.5,\rotatebox[origin=c]{270}{VAIS \citep{vais_2015}},blue) 
  \yearArrowLabel(2015.50,0.05,1.5,\rotatebox[origin=c]{90}{MarDCT \citep{mardct_2015}},teal)
  
  \yearArrowLabel(2016.50,-1.0,1.5,\rotatebox[origin=c]{270}{Modd \citep{modd_2016}},blue)
  
  \yearArrowLabel(2017.20,0.05,1.5,\rotatebox[origin=c]{90}{Marvel \citep{marvel_2016}},teal)
  \yearArrowLabel(2017.40,-1.0,1.5,\rotatebox[origin=c]{270}{SMD \citep{singapore_2017}},blue)
  \yearArrowLabel(2017.80,-1.0,1.5,\rotatebox[origin=c]{270}{Waterline \citep{waterline_2020}},olive)

  \yearArrowLabel(2018.50,-1.0,1.5,\rotatebox[origin=c]{270}{Modd2 \citep{modd2_2018}},blue)
  \yearArrowLabel(2018.80,0.05,1.5,\rotatebox[origin=c]{90}{SeaShips \citep{seaships_2018}},blue)

  \yearArrowLabel(2019.50,-1.0,1.5,\rotatebox[origin=c]{270}{MaSTr1325 \citep{mastr1325_2019}},olive)
  \yearArrowLabel(2019.70,0.05,1.5,\rotatebox[origin=c]{90}{Tamp-WS \citep{tamp-ws_2019}},olive)

  \yearArrowLabel(2020.30,-1.0,1.5,\rotatebox[origin=c]{270}{MariShipSegHEU \citep{marishipsegheu_2020}},olive)
  \yearArrowLabel(2020.50,0.05,1.5,\rotatebox[origin=c]{90}{MCShips \citep{mcships2020}},blue)

  \yearArrowLabel(2021.16,0.05,1.5,\rotatebox[origin=c]{90}{ABOShips \citep{aboships_2021}},blue)
  \yearArrowLabel(2021.18,-1.0,1.5,\rotatebox[origin=c]{270}{RoboWhaler \citep{robowhaler_2021}},black)
  \yearArrowLabel(2021.3,0.05,1.5,\rotatebox[origin=c]{90}{ROAM CRAS \citep{data_roam_cras}},black)
  \yearArrowLabel(2021.35,-1.0,1.5,\rotatebox[origin=c]{270}{DartMouth \citep{seg_lidar_cluster_jeong_2021}},blue)
  \yearArrowLabel(2021.45,0.05,1.5,\rotatebox[origin=c]{90}{USVInland \citep{usvinland_2021}},olive)
  \yearArrowLabel(2021.55,-1.0,1.5,\rotatebox[origin=c]{270}{FloW \citep{flow_2021}},blue)
  \yearArrowLabel(2021.65,0.05,1.5,\rotatebox[origin=c]{90}{GLSD \citep{glsd_2021}},blue)
  \yearArrowLabel(2021.75,-1.0,1.5,\rotatebox[origin=c]{270}{MID \citep{data_mid_2021}},blue)
  \yearArrowLabel(2021.83,0.05,1.5,\rotatebox[origin=c]{90}{WSODD \citep{data_wsodd_2021}},blue)

  \yearArrowLabel(2022.03,-1.0,1.5,\rotatebox[origin=c]{270}{MariShipInsSeg \citep{marishipinsseg_2022}},olive)
  \yearArrowLabel(2022.16,0.05,1.5,\rotatebox[origin=c]{90}{MaSTr1478 \citep{MaSTr1478_2022}},olive)
  \yearArrowLabel(2022.29,-1.0,1.5,\rotatebox[origin=c]{270}{Mods \citep{mods_2022}},blue)
  \yearArrowLabel(2022.42,0.05,1.5,\rotatebox[origin=c]{90}{Kolomverse \citep{kolomverse_2022}},blue)
  \yearArrowLabel(2022.55,-1.0,1.5,\rotatebox[origin=c]{270}{Dasha River \citep{data_dasha_river_2022}},blue)
  \yearArrowLabel(2022.68,0.05,1.5,\rotatebox[origin=c]{90}{MarPS-1395 \citep{marps-1395_2022}},olive)
  \yearArrowLabel(2022.78,-1.0,1.5,\rotatebox[origin=c]{270}{ROSEBUD \citep{ROSEBUD_2022}},olive)
  \yearArrowLabel(2022.82,0.05,1.5,\rotatebox[origin=c]{90}{FoggyShipInsseg \citep{irdclnet_2022}},olive)

  \yearArrowLabel(2023.05,-1.0,1.5,\rotatebox[origin=c]{270}{SeaSAW \citep{seasaw_2022}},blue)
  \yearArrowLabel(2023.15,0.05,1.5,\rotatebox[origin=c]{90}{Pohang Canal \citep{data_pohang_2023}},black)
  \yearArrowLabel(2023.3,-1.0,1.5,\rotatebox[origin=c]{270}{WaterScenes \citep{data_waterscenes_2023}},olive)
  \yearArrowLabel(2023.45,0.05,1.5,\rotatebox[origin=c]{90}{MariBoats \citep{mariboat_2023}},olive)
  \yearArrowLabel(2023.6,-1.0,1.5,\rotatebox[origin=c]{270}{LaRS \citep{lars_2024}},olive)
  \yearArrowLabel(2023.75,0.05,1.5,\rotatebox[origin=c]{90}{Massmind \citep{data_massmind_2023}},olive)

  \yearArrowLabel(2024.30,-1.0,1.5,\rotatebox[origin=c]{270}{MVDD13 \citep{mvdd13_2024}},blue)
  \yearArrowLabel(2024.60,0.05,1.5,\rotatebox[origin=c]{90}{Catabot \citep{data_catabot_2024}},blue)
  \yearArrowLabel(2024.90,-1.0,1.5,\rotatebox[origin=c]{270}{USV Canal \citep{data_usv_canal_2024}},blue)

     \node[anchor=south east,xshift=-300pt,yshift=-5pt] at (current bounding box.south east) 
            {
            \begin{tabular}{@{}lp{4cm}@{}}
            \ColorBox{blue} & \textcolor{blue}{object detection datasets} \\
            \ColorBox{teal} & \textcolor{teal}{classification datasets} \\
            \ColorBox{olive} & \textcolor{olive}{panoptic/semantic segmentation datasets} \\
            \ColorBox{black} & \textcolor{black}{other vision datasets} \\
            
            \end{tabular}
            };
 
\end{tikzpicture}
    \caption{Chronological overview of vision datasets for USVs.}
    \label{fig:data_chrono}
\end{figure*}
Datasets are essential for the development of computer vision models for USVs, as these models depend significantly on extensive high-quality data to achieve precise perception, navigation, and decision-making abilities in intricate marine environments. Numerous open-source datasets have been developed to address this requirement, offering varied visual information that reflects the various challenges of maritime environments, including varying weather, lighting, and sea conditions. 

In this section, we present and discuss a comprehensive range of datasets for USV vision tasks, focusing on the most commonly used tasks in this area, including object detection, classification, and segmentation (semantic, instance, and panoptic). The datasets covered in this survey span from 2015 to 2024, reflecting recent advancements and resources essential for enhancing USV vision capabilities.
Furthermore, we review the datasets that may potentially enhance the vision capabilities of USVs in the future, which provide unannotated raw data. Our objective is to engage directly with datasets gathered in real-world scenarios; consequently, in this study, we disregard all synthetic datasets that are generated through simulation or deep learning generative models. The reason for this exclusion is that the disparity in learning from synthetic data versus real-world data remains a substantial research challenge, limiting the broader applicability of synthetic data and related simulators. For each dataset, we categorize it into one of the vision tasks, such as object detection, segmentation, and others. Moreover, we rigorously analyze several characteristics of each dataset, including the number of samples provided by the dataset, the list of sensors utilized for vision, data resolution, the vision tasks for which data annotation is supported, frames per second (FPS), the number of objects pertaining to vision tasks, accessibility status, field of view (FOV), the environmental context of the USVs, such as rivers and seas, the location from which data was collected, and the year of release.

\subsection{USVs vision datasets} \label{subsec:dataset}
\subsubsection{Chronological analysis}
Figure \ref{fig:data_chrono} illustrated the chronological overview of USVs' vision datasets compiled until July 2024. Here, datasets are classified into one of three prevalent vision tasks. For the vision datasets which exclude annotations, we categorize them into other types of datasets as in the section \ref{subsec:other_data}. The earliest public dataset for USV's vision was studied in 2015 \citep{vais_2015}. In the time frame from 2015 to 2021, there were very few datasets released annually for USVs' vision. From 2021 to the present, the research on the contribution of USVs' vision datasets has experienced substantial growth. Since 2018, the popularity of object detection and segmentation tasks has grown, resulting in a decreased emphasis on classification-only datasets, as detection and segmentation provide the more detailed insights needed for advanced USV applications. As a result, the recent trend of releasing visual datasets for USVs has been primarily focussed on object detection and segmentation tasks.

\subsubsection{Object detection datasets}
Object detection datasets are essential for enhancing USV perception systems, as they enable accurate object identification and localization in dynamic maritime environments. Because USVs rely on real-time object detection to navigate safely and avoid obstacles, high-quality datasets tailored to different water surface conditions are critical for improving their autonomy and operational reliability.
Table \ref{tab:data_od} illustrates the particulars of USVs' datasets for object detection. The collection of datasets is arranged by the date of publication. In the early stage from 2015 to early 2018, the majority of datasets for object detection were concentrated on sea environments such as VAIS \citep{vais_2015}, MODD \citep{modd_2016}, SMD \citep{singapore_2017}, and MODD2 \citep{modd2_2018}. In subsequent phases, the datasets were expanded to encompass not only sea environments but also other settings including lakes, rivers, and harbors. Camera data forms the primary data source for the majority of object detection datasets. In addition to conventional cameras, infrared (IR) cameras were employed in several early datasets, exhibited by VAIS and SMD. Compared with regular camera, IR camera can enhance the ability to detect objects and obstacles in low-light or nighttime conditions, as well as through fog, smoke, or glare. Furthermore, stereo cameras, radar, as well as LiDAR and sonar have been integrated into USVs' visual datasets. MVDD13 \citep{mvdd13_2024} represents a identifiable dataset wherein panoramic camera data serves as the principal data source.
\begin{table*}[H]
\centering
\caption{Datasets for object detection in USVs. (2D OD: 2D object detction, T: object tracking, S: semantic segmentation, C: classification, N/A: lack of information.)}
\label{tab:data_od}
\resizebox{\textwidth}{!}{%
\begin{tabular}{|l|l|l|l|c|c|c|c|c|l|l|c|}
\hline
\multicolumn{1}{|c|}{\textbf{Acronym}} & \multicolumn{1}{c|}{\textbf{Number of samples}}                                                                               & \multicolumn{1}{c|}{\textbf{Sensor types}}                                                 & \multicolumn{1}{c|}{\textbf{Resolution}}                                              & \textbf{Tasks} & \textbf{FPS} & \textbf{\# objects} & \textbf{Open} & \textbf{FOV}                                       & \multicolumn{1}{c|}{\textbf{Area}}                        & \multicolumn{1}{c|}{\textbf{Location}}                           & \textbf{Year} \\ \hline
VAIS \citep{vais_2015}                                   & 2,865 images                                                                                                           & \begin{tabular}[c]{@{}l@{}}Camera\\ IR camera\end{tabular}                                                             & 1024$\times$768                                                                              & 2D OD          & N/A            & 6                  & Y            & N/A                                                  & Sea                                                          & USA                                                              & 2015          \\ \hline
Modd \citep{modd_2016}                                   & 4,454 images                                                                                                           & Camera                                                                                                                       & 640$\times$480                                                                               & 2D OD          & 10           & 2                  & Y             & 55                                                 & Sea                                                          & Slovenia                                                         & 2016          \\ \hline
SMD \citep{singapore_2017}                                    & \begin{tabular}[c]{@{}l@{}}4 on-board videos \\ (1,196 frames)\\ 32 on-shore videos \\ (16,254 frames)\end{tabular} & \begin{tabular}[c]{@{}l@{}}IR camera\\ Camera\end{tabular}         & 1920$\times$1080                                                                             & 2D OD, T       & N/A            & 10                 & Y             & N/A                                                  & Sea                                                          & Singapore                                                        & 2017          \\ \hline
Modd2 \citep{modd2_2018}                                  & 11,675 images                                                                                                          & Camera                                                                                                                     & 1278$\times$958                                                                              & 2D OD          & 10           & 2                  & Y             & 132.1                                              & Sea                                                          & Slovenia                                                         & 2018          \\ \hline
SeaShips \citep{seaships_2018}                               & 31,455 images                                                                                                          & Camera                                                                                                                      & 1920$\times$1080                                                                             & 2D OD          & 25           & 6                  & Partial       & N/A                                                  & Harbor                                                       & China                                                            & 2018          \\ \hline

MCShips \citep{mcships2020}                                  & 14,709 images                                                                                                           & Camera                                                                                                                  & Vary                                                                               & 2D OD          & N/A            & 13                  & Y             & N/A                                                   & Sea & China                                                            & 2020          \\ \hline
FloW \citep{flow_2021}                                   & 2,000 images                                                                                                           & \begin{tabular}[c]{@{}l@{}}Stereo cameras\\ Radar\end{tabular}                                                               & \begin{tabular}[c]{@{}l@{}}1280$\times$720\\ 1280$\times$640\end{tabular}                           & 2D OD          & 10           & 1                  & Y             & N/A                                                  & \begin{tabular}[c]{@{}l@{}}Lake\\ River\end{tabular}         & China                                                            & 2021          \\ \hline
GLSD \citep{glsd_2021}                                   & 212,357 images                                                                                                        & Camera                                                                                                                    & 90$\times$90 - 6509$\times$6509                                                                     & 2D OD          & N/A            & 13                 & N/A             & N/A                                                  & Port                                                         & \begin{tabular}[c]{@{}l@{}}China\\ America\\ Europe\end{tabular} & 2021          \\ \hline
DartMouth \citep{seg_lidar_cluster_jeong_2021}                                   & 493 GB                                                                                                        & \begin{tabular}[c]{@{}l@{}}LiDAR\\ Camera\\Sonar\\GPS,IMU\end{tabular}                                                                                                                      & N/A                                                                     & 2D OD, S          & N/A            & 1                 & Limited             & N/A                                                  & Lake                                                         & USA & 2021          \\ \hline
ABOShips \citep{aboships_2021}                                   & 9,880 images                                                                                                           & Camera                                                                 & 1920$\times$720                                                                              & 2D OD          & 15           & 9                  & Y             & 65                                                  & Sea                                                          & Finland                                                         & 2021          \\ \hline
MID \citep{data_mid_2021}                                   & 2,655 images                                                                                                           & Camera                                                               & 640$\times$480                                                                              & 2D OD          & N/A           & 2                  & Y             & 50                                                  & Sea                                                          & China                                                         & 2021          \\ \hline
WSODD \citep{data_wsodd_2021}                                   & 7,647 images                                                                                                           & Camera                                                               & 1920$\times$1080                                                                              & 2D OD          & N/A           & 14                  & Y             & N/A                                                  & \begin{tabular}[c]{@{}l@{}}Lake\\ River\\ Sea\end{tabular}                                                          & China                                                         & 2021          \\ \hline
Mods \citep{mods_2022}                                   & 8,175 images                                                                                                           & \begin{tabular}[c]{@{}l@{}}Stereo cameras\\ LiDAR\end{tabular}                                                             & 1278$\times$958                                                                              & 2D OD          & 10           & 3                  & Y             & N/A                                                  & Sea                                                          & Slovenia                                                         & 2022          \\ \hline
Kolomverse \citep{kolomverse_2022}                             & 215,1470 images                                                                                                        & Camera                                                                                                                      & 3840$\times$2160                                                                             & 2D OD          & 1            & 5                  & Limited       & N/A                                                  & Sea                                                          & Korea                                                            & 2022          \\ \hline

SeaSAW \citep{seasaw_2022}                                 & 1,900,000 images                                                                                                      & Camera                                                                                                                   & \begin{tabular}[c]{@{}l@{}}7680$\times$1408\\ 3840$\times$2056\\ 3648$\times$2052\\ 1920$\times$1080\end{tabular} & 2D OD, T, C    & N/A            & 12                 & N/A             & \begin{tabular}[c]{@{}c@{}}90\\ 180\end{tabular}   & \begin{tabular}[c]{@{}l@{}}Sea\\ Harbor\end{tabular}         & \begin{tabular}[c]{@{}l@{}}USA\\ Europe\end{tabular}       & 2023          \\ \hline
MVDD13 \citep{mvdd13_2024}                           & 35,474 images                                                     & Panoramic Camera                                                                                                                 & N/A              & 2D OD       & N/A           & 13                  & N/A             & N/A                                               & Sea                                                          & China                                                          & 2024          \\ \hline
Catabot \citep{data_catabot_2024}                           & 10,906 images                                                     & \begin{tabular}[c]{@{}l@{}}LiDAR\\ Camera\\ GPS,IMU\\Calibration\end{tabular}                                                                                                                   & 640$\times$480              & 2D OD,C       & N/A           & 3                  & N/A             & N/A                                               & \begin{tabular}[c]{@{}c@{}}Lake\\ Sea\end{tabular}                                                          & \begin{tabular}[c]{@{}c@{}}Korea\\ Caribbean\\USA\end{tabular}                                                          & 2024          \\ \hline
\end{tabular}
}
\end{table*}

Calibration is essential material for object detection as it ensures the accurate alignment and synchronization of data from multiple sensors, improving the precision of object localization and enhancing the overall performance of USV perception systems. However, it lack of calibration data in the public dataset of USVs. Catabot \citep{data_catabot_2024} has provided calibration; however, this dataset is not publicly available. The absence of calibration, it is challenging to exploit existing methodologies for multi-modal sensor fusion learning. In overall, the majority of object detection methodologies developed on publicly available USVs' object detection datasets are based on camera imagery.

We review the number of samples in each dataset. Certain datasets provide the unprocessed video which is utilized for the selection of frames to annotate entities, for example, SMD. In general, the quantity of images encompassed in object detection datasets ranges from thousands to millions of samples. The majority of datasets that offer an exceedingly large number of samples, such as GLSD \citep{glsd_2021} with 212,357 images, SeaSAW \citep{seasaw_2022} with 1.9 million images, Kolomverse \citep{kolomverse_2022} comprising 2,151,470 images, or DarMouth \citep{seg_lidar_cluster_jeong_2021} with 493 GB of data. Unfortunately, these large datasets are not publicly accessible. The absence of substantial public datasets may result in challenges in the development of foundational models or the benchmarking of the performance of deep learning models for object detection on a large scale in USVs.

With respect to camera image data, we consider the resolution of image data due to its significance as a pivotal factor in the development of deep learning models for object detection. Only MVDD13 and DartMouth do not explain the data resolution; all other object detection datasets specify the resolution of image data. A majority of datasets exhibit a consistent image resolution, such as VAIS, MODD, SMD, MODD2, SeaShips \citep{seaships_2018}, ABOShips \citep{aboships_2021}, MID \citep{data_mid_2021}, WSODD \citep{data_wsodd_2021}, MODS \citep{mods_2022}, Kolomverse, and Catabot. The remaining datasets encompass a mix of multiple resolution image data, including MCShips \citep{mcships2020}, FloW \citep{flow_2021}, GLSD, and SeaSAW. More specifically, several datasets integrate exceptionally high resolution image data, such as GLSD, Kolomverse, and SeaSAW, where the image resolutions are super high (\ie, 6509$\times$6509 or 7680$\times$1408). The high resolution image data depict large area scenes, making it difficult for object detection models built with lower image resolutions. Furthermore, the heterogeneity of image resolution across datasets may constitute both an advantage and a challenge in the development of object detection models for USVs, as it can facilitate the validation of models across a variety of image resolutions while also introducing a disparity of inconsistency in the development of deep learning models.

FPS information is necessary for tracking in object detection because it determines the temporal resolution with which the system processes and updates object positions. However, not all datasets present the FPS of their information. Some datasets specify the FPS during the data acquisition phase, such as the MODD, SeaShips, MODD2, FloW, ABOShips, MODS, and Kolomverse. Collectively, the FPS of these datasets is relatively low, mainly around 10 FPS. This is understandable given that USVs operate at lower speeds in the water than vehicles moving on highways.

In term of FOV aspect, this information is essential for object detection and computer vision because it defines the visible area captured by the sensor, which influences the system's ability to accurately detect objects in a scene, aids in sensor calibration, and influences the spatial context available for object detection and localization. A limited number of object detection datasets specify the FOV of their data, which include MODD, MODD2, ABOShips, MID, and SeaSAW. In general, the FOV in USVs is extensive, for example, 90, 132.1, or 180 degrees. The cameras in these discussed datasets are mounted on USVs and are primarily focused on capturing a wide scene; however, it is not always specified whether the FOV is vertical or horizontal. This indicates that the camera configuration for gathering data from USVs produce a broad FOV to enhance situational awareness of wide scene in the water environment. It is applicable to note that the water environment differs significantly from land-based transportation, wherein the roadway area is typically considerably smaller than the water expanse, and USVs require more space for manoeuvring because of their size; thus, sensors such as cameras must capture a wider scene to facilitate improved navigation of USVs. By looking at both FOV and resolution of the camera, the quality of the image can be explained. Indeed, FOV provides the extent of the scene captured, whereas resolution defines the level of detail in the image. Together, they provide a thorough evaluation of a camera's ability to balance wide-area coverage and fine detail capture for specific applications. Thus, the lack of FOV information in many datasets makes it difficult to fully understand the quality of camera image data for further analysis.

The availability of datasets is important to encourage research and development of USVs. Numerous datasets are not readily accessible (most of them are not open source), such as GLSD, Dartmouth, Kolomverse, SeaSAW, MVDD13, and Catabot. All of them are very large datasets. Ultimately, high number of objects illustrates the diversity of objects in each dataset. FloW and Dartmouth consider only vessels as target object in their annotations. The majority of datasets consider various types of objects such as ship, buoy, swimmer,... (see detail in table \ref{tab:data_object}) within the maritime environment for object detection. Specifically, several datasets introduce a diverse distinct objects within the annotation set as in table \ref{tab:data_od}, such as SMD (10 objects), MCShips (13 objects), GLSD (13 objects), WSODD (14 objects), SeaSAW (12 objects), and MVDD13 (13 objects). This indicates that annotated objects in the maritime environment for object detection exhibit significant variability and diversity. It presents a compelling challenge when developing object detection models, which need encounter the complexity of the number of objects to detect.

Aside from object detection, there are some datasets that provide additional annotations for other tasks. For example, in the SMD, it additionally facilitates the object tracking task with the annotations in the sequence of frames, while DartMouth comprises annotations of the area of objects which can be utilized for segmentation tasks. Furthermore, SeaSAW and Catabot also provide support for classification tasks. 

Overall, the development of datasets for object detection in USVs has progressed rapidly in recent years, with different type of cameras, a wide range of resolution, support for multiple vision tasks, an increase in annotated objects, and representation in multiple maritime environments and countries. It indicates that research in the domain of USVs vision is being actively encouraged.

\subsubsection{Segmentation datasets}
\begin{table*}[ht]
\centering
\caption{Datasets for segmentation in USVs. (S: semantic segmentation, P: panotpic segmentation, WS: water segmentation, N/A: lack of information.)}
\label{tab:data_seg}
\resizebox{\textwidth}{!}{%
\begin{tabular}{|l|l|l|l|c|c|c|c|c|l|l|c|}
\hline
\multicolumn{1}{|c|}{\textbf{Acronym}} & \multicolumn{1}{c|}{\textbf{\# samples}}                                                                   & \multicolumn{1}{c|}{\textbf{Sensor types}}                                                    & \multicolumn{1}{c|}{\textbf{Resolution}}                                                                                                    & \textbf{Tasks}                                         & \textbf{FPS}                                    & \textbf{\# objects} & \textbf{Open} & \textbf{FOV}                                                                            & \multicolumn{1}{c|}{\textbf{Area}}                            & \multicolumn{1}{c|}{\textbf{Location}}                    & \textbf{Year} \\ \hline
Waterline \citep{waterline_2020}                              & 400 images                                                                                                & Camera                                                                                                                     & \begin{tabular}[c]{@{}l@{}}1920$\times$1080\\ 320$\times$240\end{tabular}                                                                                 & WS                                                & \begin{tabular}[c]{@{}c@{}}60\\ 25\end{tabular} & 2                  & N/A             & N/A                                                                                       & Lake                                                          & \begin{tabular}[c]{@{}l@{}}USA\\ Switzerland\end{tabular} & 2017          \\ \hline
MaSTr1325 \citep{mastr1325_2019}                              & 1,325 images                                                                                               & Camera                                                                                                                  & 1278$\times$958                                                                                                                                    & S                                                      & 10                                              & 4                  & Y             & 132.1                                                                                   & Sea                                                           & Slovenia                                                  & 2019          \\ \hline
Tamp-WS \citep{tamp-ws_2019}                                & 600 images                                                                                                & Camera                                                                                                                     & 1920$\times$1080                                                                                                                                   & WS                                                & N/A                                               & 2                  & Y             & N/A                                                                                       & \begin{tabular}[c]{@{}l@{}}Lake\\ River\\ Harbor\end{tabular} & Finland                                                   & 2019          \\ \hline
MariShipSegHEU \citep{marishipsegheu_2020}                         & 3,560 images                                                                                               & \begin{tabular}[c]{@{}l@{}}10\% of SMD\\ 25\% of MarDCT\\ Rest: Internet\end{tabular}                                           & N/A                                                                                                                                           & C, S                                                   & N/A                                               & 2                  & Y             & N/A                                                                                       & Sea                                                           & N/A                                                         & 2020          \\ \hline
USVInland \citep{usvinland_2021}                              & \begin{tabular}[c]{@{}l@{}}29 sequences SLAM\\ 324 image pairs stereo\\ 700 images water seg\end{tabular} & \begin{tabular}[c]{@{}l@{}}LiDAR\\ Stereo camera\\ Radar\\ IMU\end{tabular}                                                 & \begin{tabular}[c]{@{}l@{}}640$\times$400\\ 1280$\times$800\end{tabular}                                                                                  & \begin{tabular}[c]{@{}c@{}}SLAM\\ WS\end{tabular} & N/A                                               & 1              & Y             & \begin{tabular}[c]{@{}c@{}}95$\times$50 (camera)\\ 360$\times$32 (LiDAR)\\ 94$\times$50 (radar)\end{tabular} & River                                                         & China                                                     & 2021          \\ \hline
MariShipInsSeg \citep{marishipinsseg_2022}                         & 4,001 images                                                                                               & \begin{tabular}[c]{@{}l@{}}5\% of SeaShip\\ 60\% from internet\\ Rest: COCO and VOC\end{tabular}                                     & N/A                                                                                                                                           & S                                                      & N/A                                               & 1                  & N/A             & N/A                                                                                       & \begin{tabular}[c]{@{}l@{}}Sea\\ River\end{tabular}           & N/A                                                         & 2022          \\ \hline
MaSTr1478 \citep{MaSTr1478_2022}                              & \begin{tabular}[c]{@{}l@{}}1,478 images\\  1,325 MaSTr1325 images\end{tabular}                            & Camera                                                                                                                        & \begin{tabular}[c]{@{}l@{}}1278$\times$958\\ 512$\times$384\end{tabular}                                                                                  & S                                                      & 10                                              & 4                  & Y             & 133.1                                                                                   & \begin{tabular}[c]{@{}l@{}}Lake\\ Sea\\ River\end{tabular}    & Slovenia                                                  & 2022          \\ \hline
MarPS-1395 \citep{marps-1395_2022}                             & 1,395 images from SMD                                                                                      & Camera                                                                                                                        & 1920$\times$1080                                                                                                                                   & P                                                      & N/A                                               & 5                  & N/A             & N/A                                                                                       & Sea                                                           & Singapore                                                 & 2022          \\ \hline
ROSEBUD \citep{ROSEBUD_2022}                               & 549 images                                                                                                & Camera                                                                                                                      & 1920$\times$1440                                                                                                                                   & S                                                      & 30                                              & 7                  & Y             & N/A                                                                                       & River                                                         & USA                                                       & 2022          \\ \hline
FoggyShipInsseg \citep{irdclnet_2022}                       & 5,739 images                                                                                               & Synthetic                                                                                                                    & N/A                                                                                                                                           & S                                                      & N/A                                               & 1                  & N/A             & N/A                                                                                       & Sea                                                           & N/A                                                         & 2022          \\ \hline
Dasha River \citep{data_dasha_river_2022}                       & 360 images                                                                                               & Camera                                                                                                                   & 1920$\times$1080                                                                                                                                           & WS                                                      & N/A                                               & 1                  & Y             & N/A                                                                                       & River                                                           & China                                                         & 2022          \\ \hline
LaRS \citep{lars_2024}                                   & 4,006 images                                                                                               & Camera                                                                                                                      & \begin{tabular}[c]{@{}l@{}}1440$\times$1920\\ 1278$\times$958\\ 1920$\times$1080\\ 640$\times$400\\ 1280$\times$720\\ 1280$\times$1024\\ 2208$\times$1242\\ 1280$\times$676\\ 1280$\times$702\end{tabular} & P                                                      & N                                               & 9                  & Y             & N/A                                                                                       & \begin{tabular}[c]{@{}l@{}}Lake\\ River\\ Sea\end{tabular}    & General                                                   & 2023          \\ \hline
MariBoats \citep{mariboat_2023}                              & 6,271 images                                                                                               & Camera                                                                                                        & N/A                                                                                                                                           & S                                                      & N/A                                               & 6                  & Y             & N/A                                                                                       & N/A                                                             & N/A                                                         & 2023          \\ \hline
WaterScenes \citep{data_waterscenes_2023}                              & 54,120 images                                                                                               & \begin{tabular}[c]{@{}l@{}}LiDAR\\ GPS\\ Radar 4D\\ IMU\end{tabular}                                                                                                       & 1920$\times$1080                                                                                                                                           & 2D OD, S, P                                                      & 30                                               & 7                  & Y             & HFOV100,VFOV60                                                                                       & \begin{tabular}[c]{@{}l@{}}River\\ Lake\end{tabular}                                                             & China                                                         & 2023          \\ \hline
Massmind \citep{data_massmind_2023}                              & 2,900 images                                                                                               & (LW)IR Camera                                                                                                     & \begin{tabular}[c]{@{}l@{}}640$\times$512\\ 320$\times$256\end{tabular}                                                                                                                                           & S                                                     & 30                                               & 7                  & Y             & HFOV24                                                                                       & \begin{tabular}[c]{@{}l@{}}River\\ Sea\end{tabular}                                                             & USA                                                         & 2023          \\ \hline
USV Canal \citep{data_usv_canal_2024}                              & 1,071 images                                                                                               & Camera                                                                                                      & 1920$\times$1080                                                                                                                                           & 2D OD, S                                                     & N/A                                               & 3                  & Y             & N/A                                                                                       & River                                                             & China                                                         & 2024          \\ \hline
\end{tabular}
}
\end{table*}

The segmentation task assigns a class label to each pixel in an image, as opposed to the object detection task's identification and localization of individual objects using bounding boxes. Segmentation is essential for USVs because it allows for precise identification and differentiation of objects in complex maritime environments, such as distinguishing between water, obstacles, and dynamic objects such as other ships. By accurately segmenting these elements, USVs can improve situational awareness, navigate safely, and make informed decisions in real time, thereby increasing operational safety and efficiency. We categorize datasets in this regard into two main types: semantic (S) and panoptic (P) segmentation datasets. Semantic segmentation datasets provide pixel-level annotations that classify each pixel into predefined categories, helping models understand the general structure of an environment, such as distinguishing water, land, or obstacles. Panoptic segmentation datasets, on the other hand, combine both semantic and instance-level annotations, enabling models to identify individual objects and their categories, which is crucial for applications like object tracking and detailed scene understanding in dynamic environments. We separated datasets that only provided water annotation into water segmentation (WS) types.

Herein, as a similar manner to the review of object detection datasets for USVs, we discuss the several attributes of segmentation datasets as presented in Table \ref{tab:data_seg}. This table shows that the development of datasets for segmentation has been studied since 2017. Waterline \citep{waterline_2020} is the first publicly available dataset for this purpose. The development has been rapidly increasing in recent years, particularly since 2022. It demonstrates that research on segmentation in USVs is gaining traction because segmentation techniques enable USVs to better detect, identify, and track objects such as other vessels, obstacles, and marine life, which is crucial for safe and efficient operation, particularly in adverse weather conditions or low visibility.

Numerous segmentation datasets are utilized for water segmentation including Waterline \citep{waterline_2020}, Tamp-WS \citep{tamp-ws_2019}, USVInland \citep{usvinland_2021}, Dasha River \citep{data_dasha_river_2022}. Several recent datasets are particularly collected for panoptic segmentation including MaSTr1478 \citep{MaSTr1478_2022}, LaRS \citep{lars_2024}, and WaterScenes \citep{data_waterscenes_2023}. The majority of datasets focus on annotating a single object (\ie, obstacle), for example, USVInland \citep{usvinland_2021}, MariShipInsSeg \citep{marishipinsseg_2022}, FoggyShipInsseg \citep{irdclnet_2022}, Dasha River \citep{data_dasha_river_2022}. Additionally, we consider the dataset that supports water segmentation which annotates solely the water area. Conversely, numerous datasets annotate a wide range of objects for segmentation such as ship, boat, buoy, swimmer,... (see detail in table \ref{tab:data_object}) including LaRS, Massmind \citep{data_massmind_2023}, WaterScenes, MariBoats \citep{mariboat_2023}, ROSEBUD \citep{ROSEBUD_2022}, MarPS-1395 \citep{marps-1395_2022}, and MaSTr1478, USV canal \citep{data_usv_canal_2024}, and MaSTr1325 \citep{mastr1325_2019}. In addition to the principal segmentation annotation, a limited number of datasets also provide annotation for auxiliary tasks, including USVInland (SLAM), MariShipSegHEU \citep{marishipsegheu_2020} (classification), as well as both WaterScenes and USV Canal (object detection).

Similar to object detection, camera data is the main sensor used for data collection related to the segmentation tasks of USVs. Several datasets provide not only camera data but also additional sensor information, such as WaterScenes (LiDAR, GPS, radar 4D, IMU), and USVInland (LiDAR, radar, IMU). Certain datasets mix multiple sources during the composing of their datasets, including MariShipInsSeg (sourced from the internet and other datasets) and MariShipSegHEU (also derived from the internet and other datasets). IR and stereo cameras are employed in a limited number of segmentation datasets. For example, Massmind utilizes an IR camera which can enhance the ability to differentiate objects based on thermal signatures, enabling effective segmentation of objects in low-visibility conditions, such as during the night or in adverse weather, where visible light cameras may struggle, whereas USVInland incorporates a stereo camera that is able to capture depth information, allowing for accurate 3D perception and object segmentation.

In terms of annotated samples, unlike USV object detection datasets, which typically contain a large number of samples, the number of samples in USV segmentation is significantly smaller, ranging from several hundreds to a few thousands. Exceptionally, the WaterScenes dataset consists of 54,120 images, which is nearly tenfold larger than the second largest datasets for segmentation. Overall, this is reasonable because the annotation task for object areas typically requires significantly more effort than the annotation effort for object bounding boxes.

Regarding the image resolution, numerous segmentation datasets are integrating samples from various origins for enhanced heterogeneity, which results in a range of image resolutions within these datasets. For example, there exist nine distinct resolutions in LaRS, and two different image resolutions in the Waterline, USVInland, MaSTr1478, and Massmind datasets. Certain datasets do not specify the image resolution of their collection, including MariShipSegHEU, MariShipInsSeg, and MariBoats. The remaining datasets describe a consistent image resolution. In comparison to object detection datasets, the image resolution in segmentation datasets is smaller and more diverse, spanning from exceedingly low resolutions such as 320$\times$240 or 320$\times$256 to 1440$\times$1920 or 2208$\times$1242. The major resolution size in segmentation datasets is 1920$\times$1080, which correlates with the standard camera resolution.

Similar to the majority of object detection datasets, there exists a limited number of datasets in segmentation that pertain to the FPS of data acquisition, which include Waterline, MaSTr1325, MaSTr1478, WaterScenes, and Massmind. In general, the FPS are consistent with the standard FPS of 30. Only the Waterline dataset provides data at an exceptionally high FPS of 60, but no video is included. Along with FPS, there are few datasets that specify the FOV of their datasets. Typically, the FOV of cameras employed in segmentation datasets remains extensive, such as 132.1 and 133.1 degrees, which is generally appropriate for capturing a broad scene in the maritime environment.

In the context of availability, there exist numerous datasets that are not publicly accessible, such as Waterline, MariShipInsSeg, and MarPS-1395. Conversely, all other datasets are publicly available. Open datasets are extremely important to encourage the research and development of USVs; thus, with the provision of open access to datasets, the development and validation of segmentation models for USVs can be accelerated.

The datasets collected for segmentation are typically heterogeneous in nature, encompassing various water environments including oceans, lakes, rivers, and harbors. Only the MariBoats dataset does not specify the collection area. This is critical for facilitating the development of models in a diverse maritime environments. Certain datasets fail to provide the specific locations of collection, such as MariShipSegHEU, MariShipInsSeg, and MariBoats. Generally, the geographical scope for the collection of segmentation datasets is confined to a limited number of countries, including USA, China, Slovenia, Singapore, Finland, and Switzerland. The absence of diverse locations may impede the validation and development of segmentation models for USVs across different geographical regions.

\subsubsection{Other datasets} \label{subsec:other_data}
\begin{table*}[ht]
\centering
\caption{Datasets for other vision tasks and unlabelled datasets. (N/A: lack of information.)}
\label{tab:data_other}
\resizebox{\textwidth}{!}{%
\begin{tabular}{|l|l|l|l|c|c|c|c|c|l|l|c|}
\hline
\multicolumn{1}{|c|}{\textbf{Acronym}} & \multicolumn{1}{c|}{\textbf{\# samples}}                         & \multicolumn{1}{c|}{\textbf{Sensor types}}                                                                & \multicolumn{1}{c|}{\textbf{Resolution}}                    & \textbf{Tasks} & \textbf{FPS}                                    & \textbf{\# objects} & \textbf{Open} & \textbf{FOV}                                           & \multicolumn{1}{c|}{\textbf{Area}} & \multicolumn{1}{c|}{\textbf{Location}} & \textbf{Year} \\ \hline
MarDCT \citep{mardct_2015}                                 & \begin{tabular}[c]{@{}l@{}}6,742 images\\ 16 videos\end{tabular} & \begin{tabular}[c]{@{}l@{}}Camera\\ Electro-optical\\ IR camera\end{tabular}                                                            & 800$\times$240                                                     & C,T            & N/A                                               & 3                  & Y             & N/A                                                      & River                              & Italy                                  & 2015          \\ \hline
Marvel \citep{marvel_2016}                                 & 1,607,190 images                                                  & Camera                                                                                                                                   & Vary                                                        & C,V,R          & N/A                                               & 25                 & Y             & N/A                                                      & Sea                                & China                                  & 2017          \\ \hline

RoboWhaler \citep{robowhaler_2021}                             & 37 videos                                                       & \begin{tabular}[c]{@{}l@{}}Center, left, right camera\\ Left, right IR camera\\ LiDAR\\ Radar\end{tabular}                                 & \begin{tabular}[c]{@{}l@{}}640$\times$512\\ 1280$\times$1024\end{tabular} & N/A              & \begin{tabular}[c]{@{}c@{}}30\\ 12\end{tabular} & N/A                  & Y             & \begin{tabular}[c]{@{}c@{}}75\\ 145\\ 360\end{tabular} & River                              & USA                                    & 2021          \\ \hline
ROAM CRAS \citep{data_roam_cras}                             & N/A                                                      & \begin{tabular}[c]{@{}l@{}}LiDAR\\ Radar\\Stereo camera\\Sonar\\Calibration\\GPS,IMU\end{tabular}                                   & 1280$\times$720 & N/A              & 60 & N/A                  & Limited             & N/A & Harbor                              & Portugal                                    & 2021          \\ \hline
Pohang Canal \citep{data_pohang_2023}                             &  N/A (1 TB)                                                       & \begin{tabular}[c]{@{}l@{}}LiDAR\\ Radar\\Stereo camera\\IR camera\\Camera\\Calibration\\GPS,IMU\end{tabular}                                  & \begin{tabular}[c]{@{}l@{}}2048$\times$1080 (S)\\ 640$\times$512(IR)\\1464$\times$2048(C)\end{tabular} & N/A              & \begin{tabular}[c]{@{}c@{}}30\\ 12\end{tabular} & N/A                  & Y             & N/A & Lake                              & Korea                                    & 2023          \\ \hline
\end{tabular}
}
\end{table*}
In addition to the datasets provided, which include annotations for segmentation and object detection tasks, we categorize the remaining datasets as in the Table \ref{tab:data_other}. This table encompasses datasets for other vision-related tasks for USVs, such as classification, object recognition, tracking, or unlabeled datasets. MarDCT \citep{mardct_2015} is the earliest released dataset for USVs that supports classification and object tracking tasks. This dataset comprises 6,742 images and 16 videos, with an image resolution of the annotated images being 800$\times$240. There are three object classes designated for each image. This dataset was collected from a river in Italy, utilizing three types of data acquisition methods: camera, electro-optical, and IR camera data. No information regarding FPS and FOV is accessible. Marvel \citep{marvel_2016} represents a substantial dataset collected from the sea environment in China in 2017, aimed at classification, verification, and recognition. The data for this dataset was acquired from various types of cameras mounted on a structure in proximity to the beach. A total of 1,607,190 images were annotated with 25 distinct classes. There is no FPS and FOV information available for this dataset. Moreover, the three remaining datasets consist of raw data devoid of any annotations. Despite their status as raw data, we retain them for analysis due to their authenticity as datasets collected within maritime environments for USVs, which may potentially be utilized for annotation in the future. These datasets are equipped with multi-modal sensor configurations for data collection. Both LiDAR, camera, and stereo camera modalities were employed in these datasets. In RoboWhaler \citep{robowhaler_2021}, the stereo IR camera was also utilized for data collection, while sonar, GPS, and IMU were employed in ROAM CRAS \citep{data_roam_cras}, and IR camera, GPS, and IMU were utilized in Pohang Canal \citep{data_pohang_2023}. Furthermore, recent datasets including ROAM CRAS and Pohang Canal also provided calibration data, which is essential for multi-modal learning approaches. All three datasets are equipped with high-resolution cameras, specifically 1280$\times$1024 for RoboWhaler, 1280$\times$720 for ROAM CRAS, and 1464$\times$2048 for Pohang Canal. These datasets also furnish information regarding FPS and are publicly accessible. While RoboWhaler was collected from a river in the USA, ROAM CRAS and Pohang Canal were gathered from the harbor in Portugal and the lake in Korea, respectively. The Small ShipInsSeg dataset \citep{inst_seg_sun_2023_danet} comprises 5,256 images with a resolution of 1280$\times$720 for instance ship segmentation annotations, but is not currently available. Campos \citep{data_catabot_2024} focuses on multi-domain inspection and maintenance of USVs. In this dataset, LiDAR, stereo camera, and sonar sensors were utilized for data collection. The diversity of maritime environments and the locations from which data were collected indicate potential for future exploitation. In summary, these datasets hold promise for the research and development of perception models for USVs in the future, particularly in the realm of multi-modal sensor fusion tasks.

\subsection{Analysis}
Based on the presentation of the vision datasets for USVs in the section \ref{subsec:dataset}, we discuss some characteristics of those datasets, such as geographical distributions, comparison with the study of vision datasets in autonomous vehicles (AVs), metadata, and the studied objects in USVs vision datasets.

\begin{figure*}[!htb]
\centering
\def\angle{0}
\def\radius{2}
\def\labelradius{4}
\def\cyclelist{{"orange","blue","gray","red","green","cyan","magenta","teal","olive","purple","pink"}}

\newcount\cyclecount \cyclecount=-1
\newcount\ind \ind=-1

\begin{subfigure}{0.48\textwidth}
      \begin{tikzpicture}[scale=0.6,font=\tiny]
        \foreach \percent/\name in {
            21.05/Not open (8),
            5.26/Korea (2),
            13.16/Slovenia (5),
            2.63/Singapore (1),
            7.89/Worldwide (3),
            5.26/Finland (2),
            13.16/USA (5),
            26.32/China (10),
            2.63/Portugal (1),
            2.63/Italy (1)
        } {
          \ifx\percent\empty\else               
            \global\advance\cyclecount by 1     
            \global\advance\ind by 1            
            \ifnum6<\cyclecount                 
              \global\cyclecount=0              
              \global\ind=0                     
            \fi
            \pgfmathparse{\cyclelist[\the\ind]} 
            \edef\color{\pgfmathresult}         
            \draw[fill={\color!50},draw={\color}] (0,0) -- (\angle:\radius)
              arc (\angle:\angle+\percent*3.6:\radius) -- cycle;
            \draw[draw=gray, shorten >=0pt] (\angle+0.5*\percent*3.6:\labelradius) node {\name~[\percent\%]} edge (\angle+0.5*\percent*3.6:\radius);
            \pgfmathparse{\angle+\percent*3.6}  
            \xdef\angle{\pgfmathresult}         
          \fi
        };
    \end{tikzpicture}
  \caption{By country}\label{fig:nation_dis}
\end{subfigure}\hfill
\begin{subfigure}{0.48\textwidth}
      \begin{tikzpicture}[scale=0.6,font=\tiny]
        \foreach \percent/\name in {
            40/Sea (16),
            30/River (12),
            20/Lake (8),
            7.5/Harbor (3),
            2.5/Unknow (1)
        } {
          \ifx\percent\empty\else               
            \global\advance\cyclecount by 1     
            \global\advance\ind by 1            
            \ifnum6<\cyclecount                 
              \global\cyclecount=0              
              \global\ind=0                     
            \fi
            \pgfmathparse{\cyclelist[\the\ind]} 
            \edef\color{\pgfmathresult}         
            \draw[fill={\color!50},draw={\color}] (0,0) -- (\angle:\radius)
              arc (\angle:\angle+\percent*3.6:\radius) -- cycle;
            \draw[draw=gray, shorten >=0pt] (\angle+0.5*\percent*3.6:\labelradius) node {\name~[\percent\%]} edge (\angle+0.5*\percent*3.6:\radius);
            \pgfmathparse{\angle+\percent*3.6}  
            \xdef\angle{\pgfmathresult}         
          \fi
        };
    \end{tikzpicture}
  \caption{By area}\label{fig:area_dis}
\end{subfigure}
\caption{Statistic of number of dataset by some characteristics.} \label{fig:geo_dis}
\end{figure*}
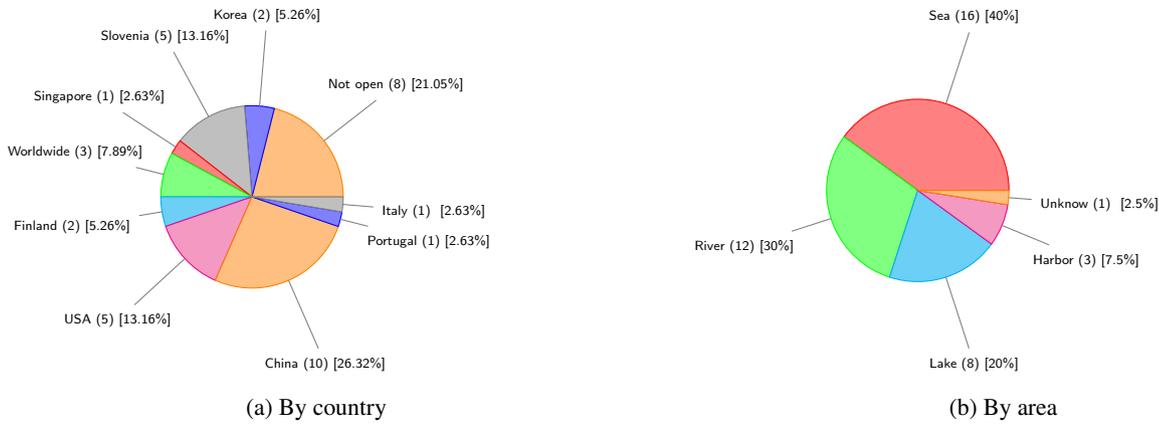

\textbf{Geographical distributions.} We analyze the global distribution of 38 vision datasets of USVs as presented in figure \ref{fig:geo_dis}. Figure \ref{fig:nation_dis} shows the distribution of datasets by country. China surpasses others with a total of 10 published datasets, while the USA and Slovenia each contribute five. A limited number of datasets were collected from other nations such as Korea, Finland, Italy, Portugal, and Singapore. Three datasets were gathered from various sea regions globally, whereas 8 remaining datasets are not accessible to the public, lacking explicit details regarding the locations from which they were collected. The number of contributed datasets from those countries signifies an extremely unbalanced development of USVs on a global scale. Similar to several existing studies that show the challenge when developing deep learning models to generalize across countries due to the heterogeneity, such as AVs \citep{av_geo_barrier,zunair2024rsud20k,app14188150} or road damage detection \citep{dg2hairoad}, USVs encounter similar challenges across different geographical regions. Nevertheless, depending exclusively on data from a singular source can cause biases that may culminate in the failure of USVs to perform adequately in diverse or unobserved regions and scenarios.
As a result, collecting data from various countries and regions is capable of addressing the particular issues that geographical locations present. This diverse regional distribution strengthens the collected data and highlights global efforts and collaborations within the research community and industry.

Furthermore, we conduct an analysis of maritime environments owing to their significance for the study of USVs. Figure \ref{fig:area_dis} illustrates the distribution of the dataset across various maritime environments. There are three main environments in maritime contexts: ocean, river, and lake. We also consider the harbor as a specialized area. We classify the implicit category as an unknown maritime environment. The statistics indicate that the majority of USVs' visual datasets are collected in the ocean with 16 datasets (40\%). Subsequently, the river follows with 12 datasets (30\%), and thereafter, the lake with 8 datasets (20\%). The harbor contributes 3 datasets (7.5\%) while the unknown environment comprises merely 1 dataset (2.5\%). This distribution is reasonable as research and development in the ocean represents the most compelling domain for USVs, followed by the riverine environment within maritime settings. Moreover, it demonstrates that the research and development of maritime technologies is not exclusively focused on the ocean, but also encompasses rivers and lakes. In terms of harbor environment, it exhibit significant variations in layout, environmental conditions, and operational practices, making data from diverse locations invaluable for improving the generalization of models in maritime applications. However, the collection and sharing of such data must carefully balance utility with privacy preservation, ensuring sensitive information about infrastructure or operations remains protected. Overall, the contributions of datasets across diverse maritime environments can further stimulate research aimed at transitioning USVs from a maritime domain to deployment across varying environments, for example, from the ocean to adaptation in rivers. It signifies the promising potential of USVs across all maritime environments in the future.

\begin{figure*}[htb]
    \centering
    \begin{subfigure}[b]{0.48\textwidth}
    \hspace{-5mm}
    \begin{tikzpicture}[label distance=.15cm,rotate=45,scale=0.15]
        \tkzKiviatDiagram[radial=3,lattice=56,gap=0.3,step=1,label space=3]%
            {2D OD,
            3D OD,
            Depth,
            3D S,
            2D S,
            OT
            }
        \tkzKiviatLine[thick,color=red,fill=red,label=SiteA](55,25,7,4,21,20);label{p4}
            \tkzKiviatLine[thick,color=blue,fill=blue](15,0,0,0,12,2);label{p5}
        
        \node [anchor=south west,xshift=-30pt,yshift=60pt] at (current bounding box.south east) {
            \tiny{
                \begin{tabular}{|l|c|c|}
                \hline
                \multicolumn{1}{|c|}{\textbf{Task}}   & \cellcolor{red!50}\textbf{AVs} & \cellcolor{blue!50}\textbf{USVs} \\ \hline
                2D OD                                 & 55          & 15           \\ \hline
                3D OD                                 & 25          & 0            \\ \hline
                OT                                    & 20          & 3            \\ \hline
                2D S                                  & 21          & 12            \\ \hline
                3D S                                  & 4           & 0            \\ \hline
                Depth                                 & 7           & 0            \\ \hline
                \multicolumn{1}{|c|}{\textbf{Sensor}} & \cellcolor{red!50}\textbf{AVs} & \cellcolor{blue!50}\textbf{USVs} \\ \hline
                LiDAR                                 & 34          & 7            \\ \hline
                Radar                                 & 19          & 6            \\ \hline
                Camera                                & 118         & 27           \\ \hline
                Stereo camera                         & 29          & 6            \\ \hline
                IR/thermal camera                     & 12          & 6            \\ \hline
                Other camera                          & 16          & 4            \\ \hline
                \end{tabular}
            } 
        };
        \end{tikzpicture}
        \caption{Vision tasks based comparison.}\label{fig:av_usv_task}
    \end{subfigure}\hfill
    \begin{subfigure}[b]{0.48\textwidth}%
    \hspace{6mm}
    \begin{tikzpicture}[label distance=.15cm,rotate=30,scale=0.14]
        
        \tkzKiviatDiagram[radial=3,lattice=120,gap=0.15,step=1,label space=5]%
            {IR/thermal camera,
            Camera,
            Radar,
            Stereo camera,
            LiDAR,
            Other camera
            }
        \tkzKiviatLine[thick,color=red,fill=red,label=SiteA](12,118,19,29,34,16);label{p4}
            \tkzKiviatLine[thick,color=blue,fill=blue](6,27,6,6,7,4);label{p5}
        \end{tikzpicture}
        \caption{Sensors based comparison.}\label{fig:av_usv_sensor}
    \end{subfigure}
    
    \caption{Comparison of number of datasets between AVs and USVs in term of vision tasks and equipped vision sensors.}
    \label{fig:av_usv}
\end{figure*}
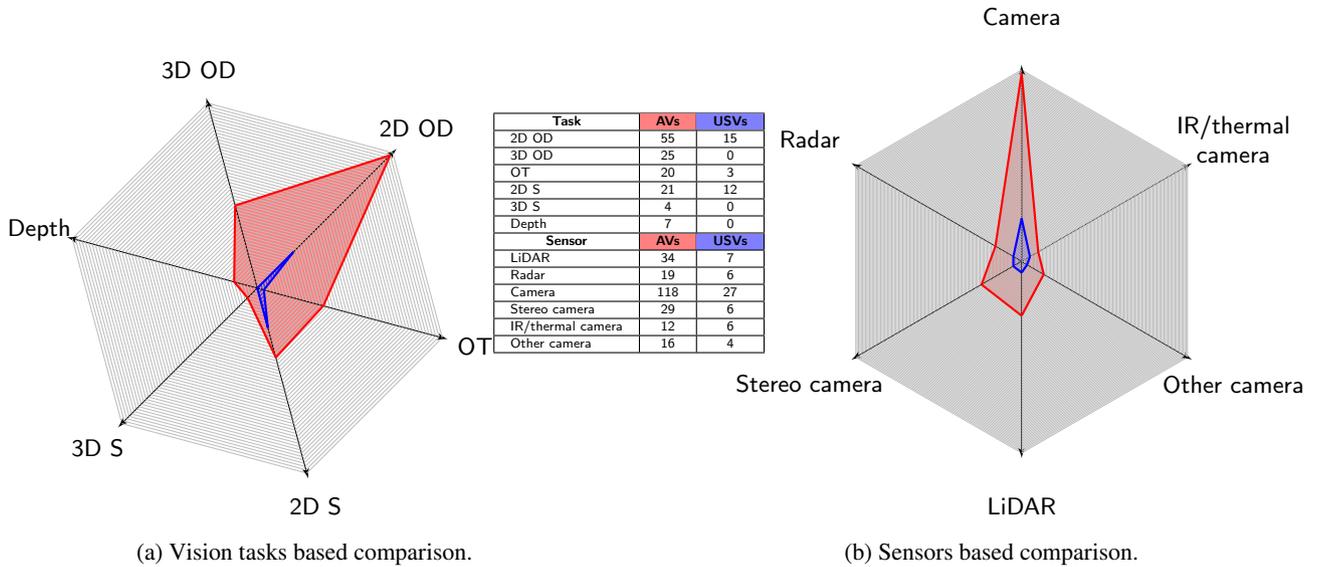
\textbf{USVs versus AVs.} AVs and USVs exhibit numerous parallels in terms of technology, functionality, and objectives. Both AVs and USVs are significantly dependent on a diverse array of sensors for perception. The most common sensors are LiDAR, radar, cameras, and ultrasonic sensors. These sensors assist the autonomous systems in detecting obstacles, estimating distances, and understanding the context surrounding them. Both categories of vehicles employ GPS and GNSS for positional tracking, along with IMU and INS paired with sophisticated mapping technologies to navigate their surroundings with precision. They leverage computer vision and artificial intelligence (AI) to analyze sensor data, detect objects, and make decisions. AI algorithms facilitate route planning, obstacle evasion, and adaptive learning from novel circumstances. Consequently, we conduct a comparative analysis between AVs and USVs with respect to the number of datasets available for vision tasks and sensors, as illustrated in Figure \ref{fig:av_usv}. For both of these analyses, we consider exclusively publicly accessible datasets. Furthermore, we account solely for the public datasets that have provided annotations for vision task-based comparisons. The studies in AVs were based on several literature reviews such as datasets \citep{av_survey_2024_dataset,av_survey_det_data_al_2024}, 3D vision \citep{av_survey_3dod_image_2024,av_survey_3dod_method_2022,av_survey_3dod_real_time_2024}, depth \citep{av_survey_depth_data_2022}, and lane detection \citep{av_survey_lane_det_2023}.

Figure \ref{fig:av_usv_task} show the comparison of AVs and USVs in the context of vision tasks. Herein, we enumerate the principal vision tasks that are pivotal for the development of AV systems. These tasks consists of 2D object detection (2D OD), 3D object detection (3D OD), object tracking (OT), 2D segmentation (2D S), 3D segmentation (3D S), and depth estimation (D). This chart demonstrates that both AVs and USVs perform certain vision tasks, such as 2D OD, OT, and S. The number of datasets supported for the 2D OD task far outnumbers those supported for other tasks in both AVs and USVs. Within the realm of AVs, 3D OD, 3D S, OT, and 2D S also attract a considerable number of contributing datasets. The number of publicly available datasets that support vision tasks in AVs is significantly higher than that of USVs. Particularly, 3D information is exceedingly critical for visual processing, including perception and navigation; however, there exists no public dataset accessible in USVs to facilitate 3D perception tasks such as 3D OD, 3D S, and D. This indicates that the development of datasets related to USVs' visual capabilities remains a challenge when compared to the development of AVs.

In addition, we also analyze the number of datasets collected by sensors for the development of vision tasks in both AVs and USVs. Sensors are crucial in supporting autonomous systems such as AVs and USVs to perceive environmental, prior to further processing by decision making systems such as planning and control. There exist several sensors utilized in both AVs’ and USVs’ systems, which comprise cameras, LiDAR, radar, stereo cameras, and IR cameras or thermal cameras. For other types of cameras, such as fisheye cameras, we categorize them into a separate group. Camera data represents the most common data type within vision datasets for both AVs and USVs due to simplicity and low cost in LiDAR environmental scenes for subsequent processing in autonomous systems. LiDAR ranks as the second most popular data type in both AVs and USVs; however, the number of datasets for LiDAR is significantly less when compared to camera-based datasets. Subsequently, radar and stereo cameras are typically employed in both AVs and USVs, whereas IR/thermal cameras and other types of cameras are less frequently utilized than the aforementioned sensors. Specifically, radars in USVs are designed for long-range navigation and detection in expansive maritime environments, while radars in AVs are optimized for short-range tasks, such as detecting nearby objects during parking or urban driving. Similar to the comparison in perception tasks, the number of datasets available for AVs across all sensors surpasses that available for USVs. This further indicates that the development of USVs' visual capabilities is receiving less attention and is less comprehensive than that of AVs, despite both autonomous systems primarily utilizing a comparable range of sensors for visual tasks. Consequently, creating robust datasets for USV vision development continues to be a greater challenge relative to AVs.

\textbf{Metadata of USVs datasets.}
\begin{table}[ht]
\centering
\caption{Available metadata of vision datasets in USVs.}
\label{tab:meta}
\resizebox{\linewidth}{!}{%
\begin{tabular}{|l|l|}
\hline
\multicolumn{1}{|c|}{\textbf{Dataset}} & \multicolumn{1}{c|}{\textbf{Metadata}}                                                                                                                                                                                                                                                                                                                              \\ \hline
SeaSAW \citep{seasaw_2022}                                & \begin{tabular}[c]{@{}l@{}}Locations: harbor, far sea shore, open sea. \\ Weather: day, dusk/dawn, night. \\ Illumination: sunny, cloudy, rain, fog, snow\end{tabular}                                                                                                                                                                                         \\ \hline
LaRS \citep{lars_2024}                                   & \begin{tabular}[c]{@{}l@{}}Environment type: River-like, sea/lake like. \\ Illumination conditions: day, dawn/dusk, night. \\ Presence of reflections: heavy, moderate, none. \\ Surface roughness: Rough, Still, Disturbed. \\ Scene conditions: Under-exposure, over-exposure, \\ sun glitter, dirty lens, fog, rain, wakes, plants/debris.\end{tabular} \\ \hline
USVInland \citep{usvinland_2021}                              & Weather: sun, rain, fog, overcast, mist, strong light                                                                                                                                                                                                                                                                                                                \\ \hline
MaSTr1325 \citep{mastr1325_2019}                              & Weather: fog, partly cloudy, overcast, sun                                                                                                                                                                                                                                                                                                                           \\ \hline
MaSTr1478 \citep{MaSTr1478_2022}                             & Weather: fog, partly cloudy, overcast, sun                                                                                                                                                                                                                                                                                                                           \\ \hline
Modd2 \citep{modd2_2018}                                  & Sudden movement, Sun glitter, land reflection                                                                                                                                                                                                                                                                                                                        \\ \hline
SeaShips \citep{seaships_2018}                               & \begin{tabular}[c]{@{}l@{}}Background\\ Lighting\\ Visible proportion\\ Occlusion\end{tabular}                                                                                                                                                                                                                                                               \\ \hline
\end{tabular}
}
\end{table}
Metadata is a significant material of datasets, particularly in autonomous systems like AVs and USVs. In the domain of AVs, a multitude of Original Equipment Manufacturers (OEMs), including Waymo, Tesla, and Alphabet, collect extensive datasets on public roadways to extract metadata that aligns with Operational Design Domains (ODDs) \citep{odd@sae} for development and validation objectives \citep{validation_odd,scenario_pdf,euro_ncap_2023,dsa4av}. Metadata provides essential details that describe a dataset's diversity and complexity. It supports the validation of deep learning models for perception tasks by clarifying the conditions under which models perform best or struggle. Consequently, dataset authors increasingly include metadata to enhance understanding and usability of datasets for model development \citep{dsa4av}. Herein, we discuss the metadata associated with vision datasets in USVs. Table \ref{tab:meta} illustrates the metadata pertaining to the vision datasets of USVs. We consider the metadata for both public and private USVs' vision datasets. This table shows that a few USV vision datasets currently include metadata when released. Common metadata information, such as location, weather, and lighting conditions, are categorized differently across datasets. Lack of metadata information or category limitations can limit the ability to represent the entire dataset for certain purposes, such as validating deep learning models \citep{dsa4av} or quantization of data for training deep learning models \citep{dataset_quantization} in USVs.


\textbf{Studied objects in USVs datasets.}
\begin{table*}[ht]
\centering
\caption{Annotated objects in the vision datasets of USVs (\cmark: annotated, -: not annotated).}
\label{tab:data_object}
\resizebox{\textwidth}{!}{%
\begin{tabular}{|l|c|c|c|c|c|c|c|c|c|c|c|c|c|c|c|c|c|c|c|c|c|c|c|c|c|c|c|c|c|c|}
\hline
\multicolumn{1}{|c|}{\textbf{Dataset}}                                  & \rotatebox{90}{\textbf{Ship/vessel/boat}} & \rotatebox{90}{\textbf{Engineering ship}} & \rotatebox{90}{\textbf{Cargo Ship}} & \rotatebox{90}{\textbf{Speed Ship}} & \rotatebox{90}{\textbf{Passenger Ship}} & \rotatebox{90}{\textbf{Float}} & \rotatebox{90}{\textbf{Sailing ships}} & \rotatebox{90}{\textbf{Marchant ships}} & \rotatebox{90}{\textbf{Fishing boat}} & \rotatebox{90}{\textbf{Official ship}} & \rotatebox{90}{\textbf{Aircraft carrier}} & \rotatebox{90}{\textbf{Carrier}} & \rotatebox{90}{\textbf{Auxiliary ship}} & \rotatebox{90}{\textbf{Container Ship}} & \rotatebox{90}{\textbf{Buoy}} & \rotatebox{90}{\textbf{Water}} & \rotatebox{90}{\textbf{River/shore bank}} & \rotatebox{90}{\textbf{Ferry}} & \rotatebox{90}{\textbf{Waste}} & \rotatebox{90}{\textbf{Person}} & \rotatebox{90}{\textbf{Swimmer}} & \rotatebox{90}{\textbf{Kayak}} & \rotatebox{90}{\textbf{Paddle/row boat}} & \rotatebox{90}{\textbf{Sky}} & \rotatebox{90}{\textbf{Bridge}} & \rotatebox{90}{\textbf{Submarines}} & \rotatebox{90}{\textbf{Missile boat}} & \rotatebox{90}{\textbf{Landing ship}} & \rotatebox{90}{\textbf{Destroyer}} & \textbf{Other Anno.}                                   \\ \hline
VAIS \citep{vais_2015}                                                                     & \cmark                               & -                         & -                   & \cmark                 & \cmark                       & -              & \cmark                    & \cmark                       & -                     & -                      & -              & -                & -               & -                       & -           & -              & -                         & -            & -              & -               & -              & -            & -                         & -             & -                & -                 & -                   & -          & -                   & -                                              \\ \hline
SMD \citep{singapore_2017}                                                                     & \cmark                               & -                         & -                   & \cmark                 & -                       & -              & \cmark                    & -                       & -                     & -                      & -              & -                & -               & -                       & \cmark           & -              & -                         & \cmark            & -              & -               & \cmark              & \cmark            & -                         & -             & -                & -                 & -      & -                   & -             & bird/plane                                              \\ \hline
Lars \citep{lars_2024}                                                                    & \cmark                               & -                         & -                   & -                   & -                       & \cmark            & -                      & -                       & -                     & -                      & -              & -                & -               & -                       & \cmark           & -              & -                         & -              & -              & -               & \cmark              & -              & \cmark                      & -             & -                & -                 & -        & -                   & -           & \begin{tabular}[c]{@{}c@{}}Animal\\ Other\end{tabular} \\ \hline
USVInland \citep{usvinland_2021}                                                               & -                                 & -                         & -                   & -                   & -                       & -              & -                      & -                       & -                     & -                      & -              & -                & -               & -                       & -             & \cmark            & -                         & -              & -              & -               & -                & -              & -                        & -            & -               & -                & -                  & -                        & -                   & -                              \\ \hline
\begin{tabular}[c]{@{}l@{}}MaSTr1325 \citep{mastr1325_2019}\\ MaSTr1478 \citep{MaSTr1478_2022}\end{tabular}          & \cmark                               & -                         & -                   & -                   & -                       & -              & -                      & -                       & -                     & -                      & -              & -                & -               & -                       & -             & \cmark            & -                         & -              & -              & -               & -                & -              & -                        & \cmark          & -                & -                & -          & -                   & -        & Void                                                   \\ \hline
ROSEBUD \citep{ROSEBUD_2022}                                                                 & \cmark                               & -                         & -                   & -                   & -                       & -              & -                      & -                       & -                     & -                      & -              & -                & -               & -                       & -             & \cmark            & \cmark                       & -              & -              & -               & -                & -              & -                        & \cmark          & \cmark             & -                & -            & -                   & -      & Flora/Debris                                           \\ \hline
\begin{tabular}[c]{@{}l@{}}Tamp-WS \citep{tamp-ws_2019}\\ Waterline \citep{waterline_2020}\end{tabular}             & -                                 & -                         & -                   & -                   & -                       & -              & -                      & -                       & -                     & -                      & -              & -                & -               & -                       & -             & \cmark            & -                         & -              & -              & -               & -                & -              & -                        & -            & -               & -                & -          & -                   & -        & Not water                                              \\ \hline
FloW \citep{flow_2021}                                                                    & -                                 & -                         & -                   & -                   & -                       & -              & -                      & -                       & -                     & -                      & -              & -                & -               & -                       & -             & -              & -                         &                & \cmark            &                 & -                & -              & -                        & -            & -               & -                & -                  & -                       & -                   & -                               \\ \hline
Mods \citep{mods_2022}                                                                    & \cmark                               & -                         & -                   & -                   & -                       & -              & -                      & -                       & -                     & -                      & -              & -                & -               & -                       & -             & -              & -                         & -              & -              & \cmark             & -                & -              & -                        & -            & -               & -                & -        & -                   & -          & Other                                                  \\ \hline
\begin{tabular}[c]{@{}l@{}}Modd \citep{modd_2016}\\ Modd2 \citep{modd2_2018}\end{tabular}                    & \cmark                               & -                         & -                   & -                   & -                       & -              & -                      & -                       & -                     & -                      & -              & -                & -               & -                       & -             & \cmark            & -                         & -              & -              & -               & -                & -              & -                        & -            & -               & -                & -                  & -                            & -                   & -                           \\ \hline
SeaShips \citep{seaships_2018}                                                                & -                                 & -                         & \cmark                 & -                    & \cmark                     & -                &  -                      & -                        & \cmark                   & -                       & -               & \cmark              & -                & \cmark                     & -              & -              & -                          & -              & -              & -               & -                & -              & -                        & -            & -               & -                & -      & -                   & -            & Bulk                                                   \\ \hline
\begin{tabular}[c]{@{}l@{}}MariShipSegHEU \citep{marishipsegheu_2020}\\ MariShipInsSeg \citep{marishipinsseg_2022}\end{tabular} & \cmark                               & -                         & -                   & -                   & -                       & -              & -                      & -                       & -                     & -                      & -              & -                & -               & -                       & -             & -              & -                         & -              & -              & -               & -                & -              & -                        & -            & -               & -                & -            & -                   & -      & Background                                             \\ \hline
MCShips \citep{mcships2020}                                                               & \cmark                                 & \cmark                      & -                 & \cmark                 & \cmark                     & -              & \cmark                     & -                       & \cmark                     & \cmark                    & \cmark              & -                & \cmark               & \cmark                       & -             & -              & -                         & -              & -              & -               & -                & -              & -                        & -            & -               & \cmark                & \cmark        & \cmark                   & \cmark          &   -                                                    \\ \hline
ABOShips \citep{aboships_2021}                                                               & \cmark                                 & -                      & \cmark                 & \cmark                 & \cmark                     & -              & \cmark                     & -                       & -                     & \cmark                    & -              & -                & -               & -                       & -             & \cmark              & -                         & \cmark              & -              & -               & -                & -              & -                        & -            & -               & -                & -         & -                   & -         &   miscellaneous                                                    \\ \hline
MariBoats \citep{mariboat_2023}                                                               & -                                 & \cmark                       & \cmark                 & \cmark                 & \cmark                     & -              & -                      & -                       & -                     & \cmark                    & -              & -                & -               & -                       & -             & -              & -                         & -              & -              & -               & -                & -              & -                        & -            & -               & -                & -                  & -                & -                   & -                                      \\ \hline

\end{tabular}
}
\end{table*}
Table \ref{tab:data_object} illustrates the maritime objects which have been annotated in the USVs' vision datasets. Here we categorize the annotations into 29 distinct objects, while alternative objects are consolidated into additional annotations. The primary objects encompass various types of vessels or boats navigating on the water surface such as general ship/vessel/boat, engineering vessel, cargo vessel, speed vessel, passenger vessel, float, sailing vessel, merchant vessel, fishing vessel, official vessel, container vessel, kayak, paddle/row boat, and some remaining naval vessels. Furthermore, several objects that are frequently observed in the water environment such as carrier, patrol, buoy, ferry, refuse, individual, swimmer, sky, and bridge are considered for analysis. This table shows that the majority of datasets provide the annotation of general ship/vessel/boat, and water objects. However, it is characterized by a notably sparse overlap of objects across datasets. Recent datasets such as MariBoats, ABOShips, MCShips, or LaRS offer a greater variety of objects compared to others. Numerous datasets that support the water segmentation task have only provided annotations of water and non-water regions. The sparse overlap of objects across datasets poses a significant challenge in the development of deep learning models for USVs. It is challenging to utilize datasets to construct transfer learning across datasets or to achieve domain generalization or adaptation learning \citep{dg2hairoad,dg_Guo@2024,dg_Faraki@2021,dg_Guo@2020,dg_Zhao@2021}. Consequently, it presents a challenge to develop deep learning models for vision tasks that encompass all entities of the USVs.

\section{Deep learning techniques}\label{sec:technique}
This section thoroughly review a wide range of methodologies studied for USVs vision. The majority of recent studies in this domain have depended on image-centric methodologies that utilize RGB visual data.
\begin{figure*}
    \begin{center}
    \resizebox{\textwidth}{!}{%
    \begin{forest}
        forked edges,
        for tree = {
            rounded corners, 
            top color=gray!5, bottom color=gray!30, 
            edge+={darkgray, line width=1pt}, 
            draw=darkgray, 
            align=center, 
            anchor=children,
            l sep=7mm,
            fork sep = 4mm
                    },
        before packing = {where n children=3{calign child=2, calign=child edge}{}},
        before typesetting nodes={where content={}{coordinate}{}},
        where level<=1{line width=2pt}{line width=1pt},
        [\textbf{USVs Vision}\\, blur shadow
            [\textbf{Single sensor}
                [\textbf{Object detection}
                    [Multiple stages]
                    [Single stage]
                ]
                    [
                        [\textbf{Others}
                            [Classification]
                            [Depth Perception]
                            [Multi-task learning]
                            [Others]
                        ]
                    ]
                [\textbf{Segmentation}
                    [Semantic]
                    [Instance]
                    [Panoptic]
                ]
            ]
            [\textbf{Multi-modal sensors}
                [\textbf{Camera-Camera}]
                    [
                        [\textbf{Camera-Lidar}]
                            [
                                [\textbf{Collaborative USVs}]
                            ]
                        [\textbf{Camera-Lidar-Radar}]
                    ]
                [\textbf{Camera-Radar}]
            ]
        ]
        \end{forest}
        }
    \caption{The taxonomy of vision techniques in USVs.}
    \label{fig:taxonomy_usv_perception}
    \end{center}
\end{figure*}
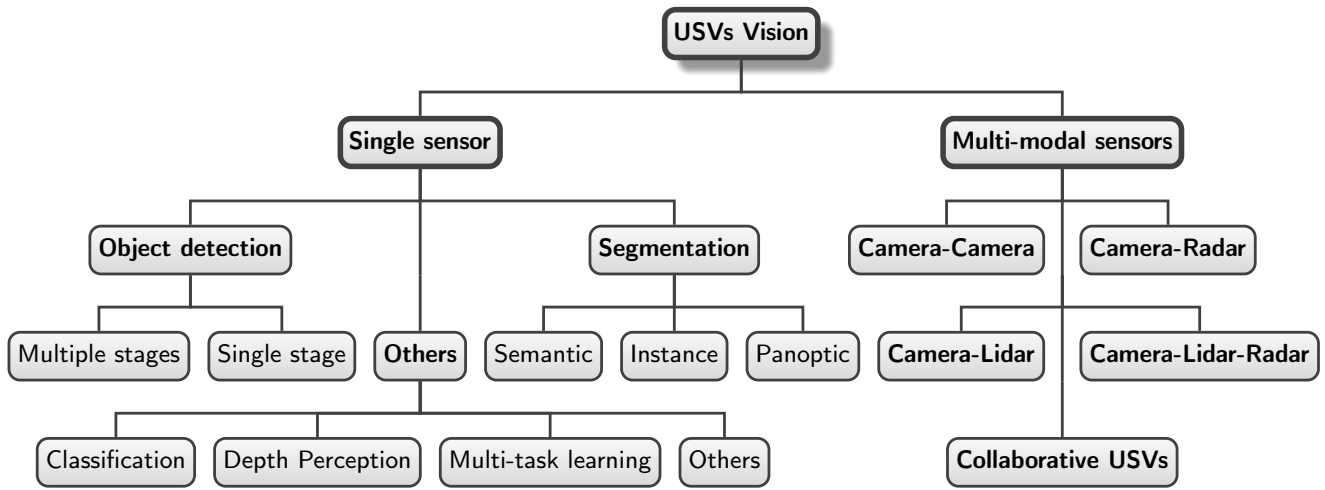
Figure \ref{fig:taxonomy_usv_perception} shows the comprehensive overview of methodologies in the vision of USVs. The deep learning techniques employed by USVs can be categorized into two principal branches: methodologies based on single sensors and those based on multi-modal sensors. The single sensor, predominantly reliant on RGB image data, consists of several important tasks such as object detection, segmentation, and additional tasks that include classification, depth perception, multi-task learning and others. We classify object detection into two primary categories: single-stage models and multi-stage models. Regarding segmentation, we concentrate on three principal tasks: semantic, instance, and panoptic segmentations. In the alternative branch of techniques, we discuss the existing methodologies concerning the fusion of multiple sensors, which include: camera-camera, camera-lidar, camera-radar, camera-lidar-radar fusion, and collaborative fusion among multiple USVs.

\subsection{Single sensor based techniques}
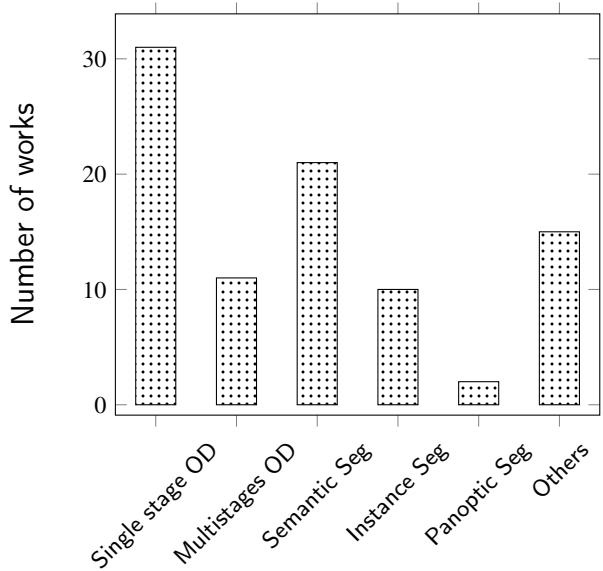
\begin{figure}[!ht]
\centering
    \begin{tikzpicture}
    \begin{axis}[
    width=0.95\linewidth,
    ybar,
    enlargelimits=0.1,
    legend style={
    legend columns=-1},
    ylabel={Number of works},
    ylabel style={font=\large},
    xtick=data,
    xticklabels={Single stage OD,Multistages OD,Semantic Seg,Instance Seg,Panoptic Seg,Others},
    xticklabel style={
      rotate=45
    },
    ]
    \addplot[ybar,bar width=15pt,pattern=dots]
    coordinates{
       (1,31)
       (2,11)
       (3,21)
       (4,10)
       (5,2)
       (6,15)
    };
    
    \end{axis}
    \end{tikzpicture}
\caption{Number of papers in single sensor based techniques for USVs. (OD: object detection, Seg: segmentation)}
\label{fig:num_single_works}
\end{figure}

Figure \ref{fig:num_single_works} shows the relative number of recent works in this topic. This chart demonstrates that there is a large amount of studies on the single-sensor based techniques, but there are still a small number of studies on panoptic segmentation due to the limitation of the number of datasets in this regard.
\subsubsection{Object detection}
Recently, the majority of object detection methodologies pertaining to maritime vision datasets rely on the existing 2D object detection techniques which have been demonstrably effective for object detection tasks on the common datasets such as Microsoft COCO, Pascal VOC,... There exist two primary approaches for object detection methodologies, which are single-stage based and multi-stage based techniques. These approaches are categorized according to the configuration of the model network architecture.

\textbf{Single stage based approach.} In this approach, the detection process is end-to-end. A substantial number of studies utilize the existing YOLO model, or merely marginally modify these models for training them on the USVs datasets. In the early stage, the object detection methodologies trained on USVs datasets comprise YOLOv2 \citep{od_yolo_usv_2018}, YOLOv3 \citep{od_yolov3_densenet_li_2020,billast2022object} or YOLOv4 \citep{od_yolov4_wang_2020,od_yolov4_kmean_he_2022}. In addition, a large number of methods based on YOLOv5 models were developed for USVs datasets \citep{od_yolov5_smd_2022,od_yolov5s_2023,od_weather_2024,od_yolov5_kang_2022,od_yolov5_kmean_zhou_2021,od_yolov5_rssd_si_2023,od_cras_yolo5s_det_cls_zhao_2023,od_yolov5s_zheng_2023,od_yolov5s_dcnn_low-light_zong_2023,od_yolov5s_optical_image_liu_2023,od_std_yolov5_ning_2024,od_yolov5_optical_cls_reg_jian_2023}. With the rapid growth of YOLO model, recent works are also based on the state-of-the-art YOLO models such as YOLOX \citep{od_yolox_transformer_ding_2023}, YOLOv7 \citep{od_yolov7_ship_jiang_2024,od_lsdnet_yolov7_lang_2024,od_yolov7_MSFFM_wu_2023,od_modan_srcnn_2020,od_yolo7_sea_2023}, YOLOv8 \citep{od_yolov8_ruiz_2023}. In addition to the YOLO family of models, numerous studies are also predicated on single stage models such as the Single Shot Detector (SSD) \citep{od_faster_rcnn_ssd_rfcn_2020,od_ssd_resnet50_li_2021,od_ssd_mobilenet_Santosa_2021,od_ssd_se_wavelet_sar_miao_2021,od_ssd_multi_scale_wen_2023,od_ssd_inland_yang_2023}, and FCOS \citep{od_fcos_he_2022,od_ofcos_zhang_2023} for training their models on USV's object detection datasets. 

In addition to employing the existing models for training with USVs object detection datasets, several studies concentrated on the modification of models to adapt with USVs' datasets. \cite{od_feature_fuse_2023} modified YOLOv7-Tiny through the implementation of multi-scale feature fusion for the enhancement of object detection precision. \cite{od_yolov8s_att_2024} developed the YOLOv8s model by integrating convolutional block attention and a self-attention paradigm. \cite{od_fe_yolo_det_2024} introduced FE-YOLO, which composes a modification of YOLOv7 for comprehensive feature fusion. The channel attention mechanism and Ghostconv of GhostNet \citep{han2020ghostnet} are utilized for network layer aggregation. A spatial pyramid pooling technique, combined with spatial channel pooling, is proposed for the coordinate attention feature pyramid network. \cite{track_yolov3_radar} proposed a detector-tracking system that employs YOLOv3 as the detection mechanism and a Kalman filter for tracking obstacles in USVs based on radar imagery.

\textbf{Multiple stages based approach.} The studies concerning this approach are mainly utilizing the recent multi-stage object detection frameworks with USVs datasets \citep{survey_ship_det_2023,survey_ship_od_2024}. Several recent models are employed to train object detection models for USVs, encompassing Faster R-CNN, \citep{od_faster_rcnn_ssd_rfcn_2020,od_weather_2024,od_faster_rcnn_2019,od_seadronesime_faster_rcnn_lin_2023}, Mask R-CNN \citep{od_mask_rcnn_2018}, FPN \citep{od_faster_rcnn_ssd_rfcn_2020,ob_isdet_maml_2023}, R-CNN \citep{od_rcnn_2019}, FPN \citep{od_faster_rcnn_ssd_rfcn_2020,od_s2anet_li_2023}, CSP with DarkNet and Attention \citep{od_radar_csp_darknet_att_yu_2023}. WaSR-T \citep{MaSTr1478_2022} introduced temporal context for the detection of water surface environment within the sequence of frames. Overall, there is relatively limited emphasis on the development of modified or enhanced network architectures within this domain of research concerning USVs.

\subsubsection{Segmentation}
This task involves the partition of images (or video frames) into multiple segments or objects. Segmentation can be thought of as a classification challenge of pixels with semantic labels (semantic segmentation), a partitioning of individual objects (instance segmentation), or a combination of the two, in which each pixel is assigned a semantic designation and each object instance is distinctly recognized (panoptic segmentation). Thus, we are primarily reviewing the deep learning techniques for three principal segmentation tasks: semantic, instance, and panoptic segmentations.

\textbf{Semantic segmentation.} Several SOTA semantic segmentation models in the COCO dataset are applied for training segmentation models on USVs datasets.  These works include UNet \citep{seg_unet_2015}, FCN \citep{seg_fcn_2015}, DeepLabv3+ \citep{seg_deeplab_v3++_2018}, BiSeNet \citep{seg_bisenet_2018}, BiSeNetv2 \citep{seg_bisenetv2_2021}, WaSR \citep{seg_wasr_2021}, IntCatchAI \citep{waterline_2020}, TMANet \citep{seg_tmanet_2021}, WODIS \citep{seg_wodis_2021}, KNet \citep{seg_knet_2021}, Segmenter \citep{seg_segmenter_2021}, SegFormer \citep{seg_segformer_2021}, STDC \citep{seg_STDC_2021}, PointRend \citep{seg_pointrend_2020}, CSANet \citep{seg_csanet_2021}.
UNet \citep{seg_unet_2015}, FCN \citep{seg_fcn_2015}, DeepLabv3+ \citep{seg_deeplab_v3++_2018}, BiSeNet \citep{seg_bisenet_2018}, BiSeNetv2 \citep{seg_bisenetv2_2021}, WaSR \citep{seg_wasr_2021}, IntCatchAI \citep{waterline_2020}, TMANet \citep{seg_tmanet_2021}, WODIS \citep{seg_wodis_2021}, KNet \citep{seg_knet_2021}, Segmenter \citep{seg_segmenter_2021}, SegFormer \citep{seg_segformer_2021}, STDC \citep{seg_STDC_2021}, PointRend \citep{seg_pointrend_2020}, WaSR-T \citep{MaSTr1478_2022}, CSANet \citep{seg_csanet_2021}.

In addition, numerous studies concentrate on tailoring the existing semantic segmentation deep learning network architectures to work with USVs datasets. \cite{seg_sem_pidnet_zhou_2024} presented PIDNet, which comprises three branches of semantic segmentation, utilizing a Row Positional Encoding Module (RPEM) and extending from PIDNet \citep{seg_pidnet_zu_2023}. \cite{seg_csp_darknet_att_multitask_cls_det_seg_yang_2024} fused CSP-DarkNet with Attention to formulate a model for multiple tasks such as classification, detection, and segmentation. \cite{seg_bev_resnet50_xu_2024} integrated a ResNet50 backbone with Transformers for bird’s eye view (BEV) segmentation which aims to provide semantically rich and spatially meaningful representations of the environment, and capture both the useful aspects of geometry and the overall layout of the scene. \cite{seg_sem_shorelinenet_yao_2021} introduced ShorelineNet, a symmetrical encoder-decoder that employ UNet with MobileNetV2 backbone. \cite{data_dasha_river_2022} adapted DCNN utilizing the ResNet50 backbone for water segmentation. \cite{seg_superpixel_xue_2021} proposed a super pixel model employing Xception and ResNet101 as backbones and devised a straightforward linear iterative clustering algorithm for water segmentation. \cite{seg_sem_exist_finl_2022} proposed the fusion of IMU data into an encoder-decoder architecture and extracted the contours. In their experiment, a combination of UNet, SegNet, and employed DeepLabV3 was used for segmentation. \cite{seg_usv_lnet_han_2023} crafted an ELNet with MobileNetV2 as a backbone and modified the decoder for segmentation. \cite{seg_lidar_cluster_jeong_2021} propose a clustering methodology to cluster lidar data into segmentations. \cite{sem_seg_knowledge_2023} employ a Bayesian framework wherein the detector is characterized by a likelihood function for the generation of probabilistic maps, and fuse outputs from this map that are derived from UNet. \cite{sem_seg_lstm_2023} propose a utilization of ConvLSTM, which integrates long short-term memory (LSTM) cells within existing convolutional neural network (CNN) structures for the purpose of semantic segmentation derived from sequential maritime videos. \cite{sem_seg_ldanet_2023} introduced LDANet, which employs a dual-branch strategy; the semantic branch implements DeepLabV3+, while the boundary refinement branch utilizes the Laplace operator to extract second-order differential features. \cite{sem_seg_visual_temp_fusion_2023} propose a video temporal feature fusion technique that integrates features from preceding frames with the current frame for the segmentation of water regions. \cite{sem_seg_eval_backbone_2022} evaluate the influence of backbone architectures on semantic segmentation in USVs, including DenseNet121, DenseNet169, DenseNet201, EfficientNet-b{0-7}, Inceptionv3, MobileNet, MobileNetv2, SeNet154, and InceptionResNet-v2. \cite{sem_seg_fuse_cityscape_2023} propose a mask fusion approach between WaSR-T and Panoptic DeepLab for the water segmentation of USV images. \cite{sem_seg_attentropy_2022} propose a network which consists of various attention layers of vision transformers in the training of semantic segmentation for USV data. The extraction of entropy from spatial attentions is proposed. \cite{sem_seg_low_light_2021} propose an adaptation of UNet that incorporates an attention mechanism and dense connectivity to augment segmentation performance in low-light environments. \cite{sem_seg_sgaf_2022} present SGAF, which employs MobileNetV2 as a foundational backbone, and propose a network architecture featuring aligned fusion and utilizing ground truth for enhanced guidance in water segmentation. \cite{sem_seg_swinseg_2023} introduced SwinSeg, a hybrid network that fuses Swin Transformer and a lightweight multi-layer perceptron (MLP) for ship segmentation. \cite{od_wodis_2021} proposed WODIS, an encoder-decoder network designed for semantic segmentation. The network comprises an attention refinement module and a feature fusion module that concatenates multidimensional features within the decoder network.

\textbf{Instance segmentation.} In contrast to semantic segmentation, which labels all objects of the same class as one, instance segmentation assigns a class label to each pixel while also distinguishing between individual objects within the same class, allowing for precise object separation. This is critical in applications that require the identification of unique instances, such as counting objects or tracking specific individuals. \cite{marishipsegheu_2020} introduced IFDSE, which comprises two phases: the Interference Factor Discriminator (IFD) founded on SqueezeNet \citep{inst_seg_squeezenet_2016} and the Ship Extractor (SE) derived from DeepLabv3+ \citep{inst_seg_deeplabv3+}. This methodology aims to categorize maritime scenes into fog-affected and fog-free scenarios, subsequently facilitating the extraction of vessels. \cite{inst_seg_fog_scene_2021} proposed a channel attention module to address the challenges associated with ship instance segmentation within fog-laden environments. \cite{inst_seg_sun_2023_danet} presented DANet, which encompasses three principal components: Feature Encoding, a Dual Mask Branch, and a Dual Activation Branch. The Feature Encoding employs a pyramid pooling module alongside a feature refinement module. Both the Dual Mask Branch and Dual Activation Branch are composed of a succession of convolutional blocks, culminating in the fusion of outputs from these two components. The proposed training loss integrates dice loss and cross-entropy loss. \cite{inst_seg_massnet_2024} introduced MASSNet, featuring a multiscale attention backbone that leverages attention mechanisms and triplet attention \citep{inst_seg_triple_att_2021} to augment multiscale feature extraction across diverse dimensions. The proposed training loss incorporates focal loss for object classification and dice loss for mask prediction. \cite{inst_seg_ma_2023_mrisnet} presented MrisNet, which utilizes FasterYOLO, a derivative of YOLOV5s for feature extraction, and integrates it with a Transformer architecture for multiple heads, serving both segmentation and detection of vessels. \cite{inst_seg_mdd_shipnet_2024} proposed the MDD-ShipNet architecture, which is constituted by polarized self-attention (PSA) and a weighted bidirectional feature pyramid network (WBiFPN) for detection, as well as five filters based on CNN for defogging purposes. \cite{inst_seg_sun_2023_cascade} introduced CascadeAgg, an aggregation segmentation network that employs multi-scale edge aggregation information. The GIoU loss was employed in lieu of IoU loss for the training of the model.

\textbf{Panoptic segmentation.} This technique combines the benefits of semantic and instance segmentation by labeling each pixel with a class and a unique object identification, allowing it to distinguish between individual objects and background classes in a single, unified output. This is especially useful for comprehensive scene understanding because it includes both object boundaries and instance-level details. Along with the limitations of the number of datasets for panoptic segmentation in USVs, the studies of this task remain markedly insufficient. Numerous baseline models that have demonstrated efficacy in general panoptic datasets are employed to train on the most recent USVs panoptic dataset, \ie, LaRS. These include Panoptic-Deeplab \citep{pan_seg_deeplab_2020}, Panoptic-FPN \citep{pan_seg_fpn_2019}, and Mask2Former \citep{pan_seg_mask2former_2022}. In addition to that, recently, ICLRNet \citep{irdclnet_2022} is proposed upon an anchor-free Fully Convolutional One-Stage (FCOS) architecture with five heads and integrates Dynamic Contour Learning, which encompasses dynamic convolutions designed for delineating indistinct vessels in obscured conditions by obtaining the comprehensive contour.

\subsubsection{Others}
Aside from object detection and segmentation, a few studies have focused on developing other vision tasks in USVs, such as depth perception, multi-task learning, classification, and other tasks.

\textbf{Depth perception.} This is crucial for USVs as it enables accurate distance measurement to objects and obstacles, which is essential for safe navigation, collision avoidance, and precise maneuvering in dynamic maritime environments. \cite{fuse_audio_cam_2024} fuses auditory and visual textual images in a maritime environment for the recognition and localization of vessel targets. The Symbiotic Transformer, which is constituted by cross-modal attention for vessel target recognition. A heterogeneous batch normalization layer is employed for addressing conflicting gradients among disparate modalities. Here, multiview regression is utilized for target depth prediction. \cite{depth_wdnet_2023} introduce WDNet, a depth estimation models for the obstacles in the water surfaces. The network employs a Bottleneck in ResNet18, along with a decoder comprising a cascade of multiple Atrous Spatial Pyramid Pooling (ASPP) blocks. The synthetic depth is derived from binocular stereo video sequences for assessment due to the absence of depth ground truth.

\textbf{Multi-tasks learning.} Unlike with single task learning, this technique allows a single model to perform multiple tasks (such as detection, classification, and segmentation) simultaneously, optimizing computational efficiency and improving the accuracy of each task by leveraging shared representations. A few of studies concentrated on training models across various vision tasks for USVs. \cite{multitask_2023} propose a multi-task learning framework encompassing vessel detection, water segmentation, and camera quality classification for the visual perception of water surfaces. This research utilizes QARepVGG \citep{sem_seg_QARepVGG_2024} as the foundational architecture, FPN \citep{od_fpn_2017} for subsequent processing, and devises a cascade of three heads. Overall, the research concerning multi-task learning within the visual domain of USVs remains relatively constrained.

\textbf{Classification.} This is a typical task for USVs. \cite{cls_transformer_few_shot_2023} propose a classification model predicated on Transformer with augmented self-attention for the classification of vessels within maritime environments. \cite{cls_adaptive_learning_xu_2021} introduced an adaptively used recurrent memory network that selectively retains instances that are challenging to identify. \cite{cls_contour_Chen_2020,cls_cascade_cnn_Chen_2020} developed a multilevel cascaded convolutional neural network transitioning from coarse to fine classification. \cite{cls_comparison_cnn_yao_2019,cls_fu_2018} proposed a cascaded convolutional neural network grounded in shape and geometric principles to enhance the edge detection algorithm employing an adaptive tuning kernel technique to construct a versatile feature encoder. \cite{cls_multiscale_Tian_2023} propose a multiscale and multilevel feature network which utilizes ResNet50 as the foundation for vessel target recognition in complex maritime environments.

\textbf{Others.} Aside from the techniques mentioned above, several other techniques have been studied to improve USV vision. \cite{calibaration_stereo} proposed calibration utilizing a checkerboard pattern derived from stereo images for USVs. \cite{skye_line_gradient_2020} introduced a method for detecting the sea-sky line based on gradient saliency. \cite{fusion_3d_ship_reconstruction_2024} present a Generative Adversarial Network (GAN)-based three-dimensional ship reconstruction from singular view images. The architecture comprises several processes, including depth and albedo feature extraction prior to the generation of the three-dimensional morphology of the vessel. \cite{seadsc} proposed the dynamic scene change detection within maritime camera recordings with the aid of latent feature extraction and similarity assessment frameworks.

\subsection{Multi-modal sensors based techniques}
\begin{figure}[!ht]
\centering
    \begin{tikzpicture}
    \begin{axis}[
    width=0.95\linewidth,
    ybar,
    enlargelimits=0.1,
    legend style={
    legend columns=-1},
    ylabel={Number of works},
    ylabel style={font=\large},
    xtick=data,
    xticklabels={camera-camera,camera-lidar,camera-lidar,camera-radar-lidar,collaborative USVs},
    xticklabel style={
      rotate=45
    },
    ]
    \addplot[ybar,bar width=15pt,darkgray!20!black,fill=darkgray!80!white]
    coordinates{
       (1,9)
       (2,10)
       (3,13)
       (4,4)
       (5,2)
    };
    
    \end{axis}
    \end{tikzpicture}
\caption{Number of papers in multimodal sensor fusion techniques for USVs.}
\label{fig:num_fusion_works}
\end{figure}
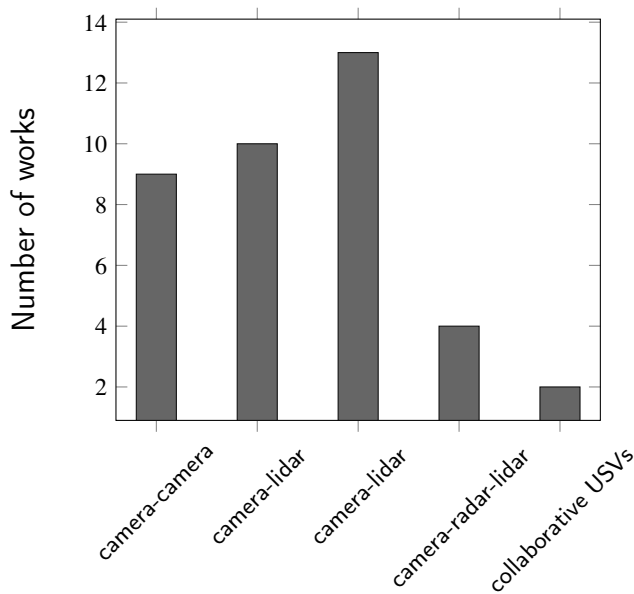

In addition to single sensor-based vision methodologies for USVs, numerous approaches have recently focused on the aggregation of multiple data sources to augment the efficacy of a diverse array of vision tasks in USVs. Such studies are designed to exploit the heterogeneous attributes of multiple sources and construct a unified, coherent, and richer representation that can enhance superior situational awareness. Figure \ref{fig:num_fusion_works} shows the relative number of recent works in this topic. This chart demonstrates that there is a significant amount of research on multimodal sensor fusions, but there is still a small number of studies on collaborative USV fusion. 
      
      

\subsubsection{Camera-camera} 
The fusion of multiple camera inputs (camera-camera) in USVs can improve spatial coverage, depth perception, and resistance to visual occlusions, thereby improving obstacle detection, navigation accuracy, and overall situational awareness in challenging marine environments. \cite{fusion_tracking_2023,fusion_heterogeneous_2022} propose camera-based tracking method by employing clustering techniques for multi-camera fusion. YOLOv4 was utilized for the detection task of each individual camera. Subsequently, the anticipated bounding boxes are consolidated for fusion predicated on pseudo point cloud projection in conjunction with a calibration matrix relative to one another. \cite{fusion_cam_lidar_radar_2020} propose a data fusion from RGB cameras and IR cameras, which integrates three channels of RGB imagery and one channel of IR imagery into four channels to enhance object detection in USVs. \cite{fuse_stereo_od_2024} propose obstacle detection for USVs utilizing stereo cameras. The procedure encompasses three stages: sea-sky line detection via Hough transformation, object detection on each camera utilizing Single Shot MultiBox Detector (SSD), and feature matching along with point cloud generation for binocular imagery. \cite{fusion_track_3d_lidar_2019} propose an algorithm for object detection on the water surface through approximate semantic segmentation. The three-dimensional point cloud of the area in front of the vessel is extracted from the stereo camera. A histogram-like depth appearance feature is employed to track objects across adjacent frames.
It is the hierarchical orthographic projection perception segmentation, also referred to as a BEV \citep{data_WSBEV_2024}. This research introduces a hybrid BEV segmentation network that employs the shared backbone and geometry-based transformer for view transformation of hybrid cameras. \cite{track_multi_vessels} propose a synthesis of Electro-Optical and IR camera data, and an Augmented Extended Kalman Filter (AEKF) was developed which extends the Extended Kalman Filter (EKF) \citep{track_ekf} for tracking filtration with the foundational detector being YOLOv5. \cite{track_multi_object} propose a tracking mechanism utilizing the YOLOv7 detector and an observation-centric Kalman filter that exploits motion between consecutive frames in video to improve the performance of obstacle detection. \cite{track_stereo_point_cloud} present an object tracking algorithm for monitoring obstacles in USVs through a virtual point cloud generated from a stereo camera. In this system, the identified obstacles are projected onto 2D-bounding boxes from the virtual point cloud produced by the stereo camera and IMU, and then temporal validation was used to filter out false detections. 

\subsubsection{Camera-LiDAR} 
The fusion of LiDAR and camera (Camera-LiDAR) in USVs help improving of object detection, distance estimation, and navigation in a variety of marine conditions and environments. \cite{fusion_small_od_2021} proposed a fusion architecture for camera and LiDAR data to detect small objects from USVs. The sequence of LiDAR frames is consolidated to serve as input to a backbone for feature extraction. Subsequently, the derived features from the images are fused with those features utilizing self-attention and global attention mechanisms. Ultimately, a FPN prediction with three heads is executed to identify small, medium, and large objects. \cite{fusion_transformer_distillation_2023} presented the fusion of LiDAR point clouds and camera data for water segmentation. The network employs a Transformer for the fusion of 3D point clouds and 2D images, augmented by knowledge distillation to produce 2D images from 3D point clouds. \cite{fusion_seg_water_cam_lidar_2022} propose cross-modal fusion between camera and lidar data for water segmentation. The 3D point clouds are supplemented with 2D images. The outputs of 3D segmentation are directed into a subsequent 2D segmentation model for 2D water segmentation. \cite{seg_pointnet++_qi_2017} utilized PointNet++ for 3D feature extraction, while UNet \citep{seg_unet_2015} is employed for 2D segmentation. This methodology was validated on a private data collected in China. \cite{fusion_multi_modality_2020} propose a neural network that fuses RGB images with multiple point clouds derived from LiDAR and depth cameras. The calibrated images and point clouds are subsequently inputted into the network for object detection, and 3D detection is conducted through a generative network. The 3D proposal box is projected onto the BEV and image feature map for depth fusion detection networks to acquire 3D objects. The method was validated on a self-collected dataset. \cite{fusion_fly_boat_2023} initially employ YOLOv7 for the detection of 2D objects from images, then project LiDAR point clouds onto images using a calibration matrix, and ultimately extract the distance and angle of the maritime vessel. \cite{fusion_cam_lidar_zhang_2021} propose feature fusion predicated on camera and LiDAR data. Initially, the camera undergoes preprocessing with sea-sky separation to eliminate spurious objects. The point cloud is projected onto the image via a calibration matrix. Finally, YOLOv3 is utilized for the extraction of candidate regions from the fused image. \cite{fusion_multisensor_2023} proposed camera and LiDAR fusion for 2D object detection. Initially, DBSCAN is employed in point clouds for candidate extraction, while YOLOX is applied to camera data for bounding box extraction; thereafter, the Regions of Interest (RoI) from these two outputs are fused. Tracking component with the bounding box from the preceding frame is implemented during the fusion process. Validation is conducted using a private dataset. \cite{fusion_lidar_cam_3dod_2024} propose multi-modal information fusion for object detection. The framework encompasses three components: pre-fusion with target Regions of Interest (ROIs) formed by point clouds (combined SVM and DBSCAN) and image data (Faster R-CNN) respectively, fusion of ROIs, and post-fusion predicated on 3D bounding boxes and 2D bounding boxes. This method was validated on a private dataset in Tianjin, China.

\subsubsection{Camera-radar} 
The fusion of camera and radar (camera-radar) in USVs improves robustness to adverse weather and low-visibility conditions because radar can detect objects through fog, rain, or darkness, complementing the camera's visual detail-an advantage over camera-camera or camera-LiDAR fusion, which are more limited under such conditions. Moreover, RADAR in the USVs is its capability to detect objects over longer distances, which does not happen in the case of LiDAR. \cite{fusion_realtime_cam_radar_2023} introduced a network referred to as FVMNet for real time lidar point cloud generation from the camera and radar of USVs. The architecture comprises an encoder-decoder structure, followed by a spatial transformer to transform image features into BEV space, along with the fusion segmentation head on top of the network. \cite{fusion_efficient_vrnet_2023} propose a contextual clustering and asymmetrical fusion of 4D radar and camera for multiple tasks, including object detection and segmentation of USVs. A sequence of radar frames is utilized for the fusion of local radar features. \cite{fusion_mask_vrdet_2024} introduced Mask-VRDet, which integrates camera and 4D radar for semantic segmentation of USVs. Initially, extraneous points were eliminated using PointNet++ \citep{seg_pointnet++_qi_2017} and DBSCAN. Subsequently, a graph neural network is employed to merge features from the image and radar point cloud to produce a segmentation map. \cite{fusion_radar_cam_2021} conduct late fusion by utilizing correlations between sensor measurements for vessel detection in radar and camera imagery. YOLOv3 is utilized for the detection phase in both camera and radar. RV-YOLOX \citep{fusion_rv_yolox_2024} first converts the radar's 3D point cloud into a 2D image via a modification of UNet integrated with a Point Transformer network, and thereafter proposes radar-camera fusion object detection based on YOLOX with a spatial attention fusion block in the decoder. \cite{fusion_small_det_rad_cam_2024} introduced a network named RCFNet for the detection of small objects through the fusion of radar and camera data. The network employs a sequence of radar frames for feature fusion and utilizes image-guided radar features by integrating semantic embedding information, along with a cross-transformer for radar and range-Doppler fusion. MobileVIT \citep{fusion_mobilevit_2021} and RCNet \citep{fusion_achelous_2023} serve as the foundational architectures for RGB and radar data, respectively, for an object detection model in USVs. Achelous \citep{fusion_achelous_2023} proposed a framework for the fusion of camera and radar aimed at USVs perception tasks. This encompasses five tasks: detection, segmentation of objects, movable-area segmentation, waterline segmentation, and radar point cloud segmentation. Vision Transformer (VIT)-based models such as EdgeVIT \citep{fusion_edgevits_2022} and MobileVIT \citep{fusion_mobilevit_2021} are employed for image-based feature extraction, while RCNet functions as the encoder for radar maps. PointNet++ \citep{seg_pointnet++_qi_2017} was utilized for the Point Cloud Semantic Segmentation model. In common, since most of the vessels are already equipped with radars, therefore it is cost-effective to use radar for long distance measurements in maritime environments.

\subsubsection{Camera-lidar-radar} 
The fusion of LiDAR, radar, and camera in USVs provides comprehensive environmental perception by combining radar's robustness in adverse conditions, LiDAR's accurate depth data, and the camera's visual detail, resulting in superior object detection, situational awareness, and navigation when compared to dual-sensor fusion approaches such as camera-camera, camera-LiDAR, or camera-radar.
\cite{fusion_multisensor_2018} propose a multi-sensor fusion method for marine object detection encompassing radar, lidar, RGB, and infrared cameras. The subsequent fusion derived from the detection outcomes was determined through probabilistic data association. \cite{fusion_few_shot_2022} introduce three-dimensional object detection for USVs utilizing camera, lidar, and radar. ResNet50 is employed for feature extraction; a region proposal network (RPN) is utilized for the integration of features across modalities, and few-shot learning predicated on metric learning is proposed for label classification. \cite{fusion_cam_radar_lidar_2021} present a software module and algorithm for the object detection and tracking of USVs utilizing radar, lidar, and three monocular cameras. The object detection and segmentation are initially undertaken on the camera, subsequently, the output is fused with radar and lidar filtering for the ultimate obstacle tracking. \cite{fusion_depster_shafer_2022} propose the application of Dempster-Shafer (D-S) \citep{fusion_dempster_shafer_1976} evidence theory for fusion to identify objects through a four-step process: a grid map of the environmental model of USVs, sensor data from three-dimensional lidar, radar, and stereo cameras were employed as evidence for the attributes of each grid, detection from each sensor, and the D-S combination rule for the final fusion decision regarding grid attributes. An experiment was conducted utilizing private data collected in China. Overall, due to a lack of public datasets in the sensor fusion branch for USVs, the studied methods remain limited.

\subsubsection{Collaborative USVs} 
Collaborative (multiple) USVs fusion improves environmental perception, task efficiency, and resilience by allowing USVs to share sensor data and insights in real time, resulting in coordinated decision-making, increased operational range, and improved accuracy in complex or dynamic marine environments. \cite{fusion_coastal_color} execute spatio-temporal calibration of multiple sensors (cameras, lidar) for dual USVs concerning object and background in dual-view based lidar. Feature-based fusion is employed. The background information is utilized to rectify illumination discrepancies in dual view. \cite{fusion_unsupervised_dual_view_lidar} introduced an unsupervised minor obstacle detection methodology from dual USVs. The framework utilizes the Siamese graph neural network for global contour feature extraction, and a rotation matrices estimation network to ascertain rotation transformation for aligning point clouds. Gridding, filtering, and clustering are employed to ascertain the dimensions and locations of obstacles. An experiment was conducted on a self-assembled dataset named QDcoast. In general, research in this field remains limited due to challenges such as the complexity of real-time data integration and communication among multiple USVs, which can be hindered by latency, heterogeneous systems, or environmental factors such as signal interference.

\section{Challenges and future perspectives}\label{sec:discussion}
Notwithstanding the comprehensive efforts and progress accomplished in the domain of USVs vision datasets and deep learning methodologies, there are numerous unresolved challenges and prospective research trajectories that need to be addressed in order to achieve further progress.
\subsection{Lack of 3D vision datasets} 
The navigation of USV systems requires 3D information. Several principal 3D vision tasks, such as depth estimation, 3D object detection, and 3D segmentation, are formulated to acquire 3D information. The lack of public 3D vision dataset in USV makes it challenging to develop and validate deep learning models for 3D vision perception. As previously mentioned, a limited number of public datasets have offered multi-modal sensors; however, it lack of annotations on these sensors, and there was almost no calibration data supplied. In the absence of calibration data, the development of deep learning models for USV's vision tasks predicated on multi-modal sensors will be difficult to progress, presenting a substantial challenge for 3D information acquisition in USV. 

\subsection{Sparsity of annotated objects} 
The maritime environment encompasses a variety of moving objects, including animals, vessels, and watercraft, which can exert a direct influence on USVs. Subsequently, detection of these objects are essential for the visual perception of USVs. Nevertheless, currently available datasets still comprise a limited quantity of annotated objects. To ensure the ability to capture a wide range of dynamic objects on the water surface, the process of data acquisition necessitates an extended duration and could be conducted across a multitude of locales, conditions, and so forth. Therefore, future research on contributing vision datasets for USVs should take into account the diversity and quantity of objects.

\subsection{Inconsistency of objects across datasets} 
The lack of overlap between datasets makes challenging the process of developing and validating vision models' efficacy on unobserved domains, which is influenced by domain adaptation and domain generalization learning. Even if multiple datasets contain the same object, such as a ship/vessel/boat, different objects from each dataset will require standardization before achieving object consistency; however, very rarely overlapping objects may be preserved for future use. Domain generalization and adaptation are current trends in the development of vision models for autonomous systems such as AVs and USVs to reduce the disparity of unobserved data; thus, potential research in USV vision may focus on contributing datasets with the same objects in addition to developing domain generalization and adaptation models.

\subsection{Metadata of USV's datasets} 
A limited number of datasets provided metadata information regarding their dataset. Metadata is now an essential asset for conducting research and developing deep learning architectures for vision tasks. Therefore, by employing metadata, one can quantify datasets derived from large datasets for objectives such as condensing datasets to accommodate storage constraints, curating qualified data for annotation purposes, and selecting data for the training of deep learning models that maintain performance comparable to that of the entire dataset \citep{dsa4av}, even when hardware resources such as GPUs are limited. Furthermore, metadata may be employed to verify and identify the scenarios or conditions under which models or autonomous systems, such as AVs and USVs, perform satisfactorily or inadequately, allowing for future enhancement. In terms of USVs, one potential development direction is to focus on metadata aggregation and metadata-informed deep learning models.

\subsection{Large field of view, but small objects} 
Owing to GPU resource limitations, general object detection and segmentation frameworks are formulated with diminished image resolutions, such as $640\times640$, $800\times800$, or $1024\times1024$ in YOLO-family architectures. Nevertheless, the image resolution in USV vision datasets is typically high (thousands to thousands of pixels). The objects appear extremely small due to the large field of view used in USV data collection. So, with this issue, deep learning models require being developed at high resolutions and with significant foreground-background disparities. Extensive images are typically handled by resizing them to reduced dimensions, which can result in the object appearing so small in the resized image that the model is unable to recognize it. Furthermore, the camera FOV in USV's vision datasets is wide, allowing for the capture of a larger area surrounding the USVs. This may lead to an increase in the number of objects in the dataset. As a result, it is essential to allocate resources to the development of deep learning models to deal with these issues for USVs.

\subsection{Lack of incorporation of LLMs} 
Large Language Models (LLMs) have recently seen remarkable applications across a wide range of domains. Notwithstanding the similarity between AVs and USVs, many recent studies have concentrated on developing LLM frameworks for AVs \citep{llm_av_1_survey}. For example, LLM is employed to provide linguistic information to visual tasks in order to improve explanation and reliability. LLM frameworks are additionally utilized to produce data for autonomous navigation, such as imagery or video descriptions, event plausibility, or generating video or image content predicated on a description. Nevertheless, there exists a dearth of research regarding the integration of LLM into USVs. For example, several data information sources, such as meteorological data, including wind, wave, and current patterns, can be useful for the development of LLM models for USVs. Therefore, future research into USVs should concentrate on integrating the LLM framework to USVs in order to enhance USV functionality, particularly in the realm of USV vision.

\subsection{Data Privacy} 
Maintaining privacy in AVs data constitutes a paramount concern to tackle \citep{pp4av,fisheyepp4av}. As machine learning has become increasingly prevalent in AVs, corporations and research organizations collected a substantial volumes of data for development and validation purposes. More collected data entails greater responsibility for data privacy. Many regulations from the European GDPR~\citep{gdprEU}, California CCPA~\citep{ccpaCali}, Chinese CSL~\citep{cslChina}, or Japanese APPI~\citep{appiJapan} necessitate that participants' personally identifiable information be protected and removed upon request. In reaction to these regulations, numerous commercial solutions have been formulated to anonymize collected data (predominantly through the blurring of visual data).
\begin{figure}
    \centering
    \includegraphics[width=0.9\columnwidth]{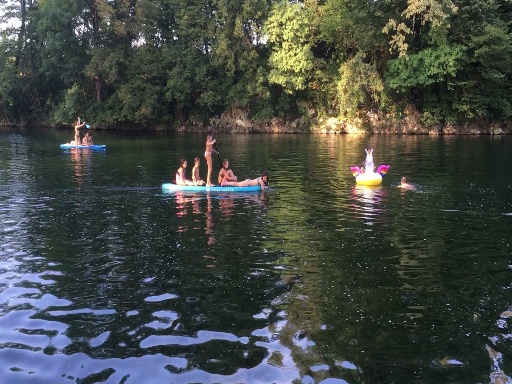}
    \caption{A sample in MaSTr1478 dataset \citep{MaSTr1478_2022} without data privacy.}
    \label{fig:dis_privacy}
\end{figure}
However, there has been little research into data privacy in USVs. Figure \ref{fig:dis_privacy} illustrates an example image in MaSTr1478 \citep{MaSTr1478_2022}, in which the face of a human was leaked, resulting in a potential violation of regulations. Among public datasets, only LaRS \citep{lars_2024} claimed that human identity was blurred before publishing their dataset, but no explicit method of data privacy was explained. Hence, future research should focus on data privacy techniques as well as a dataset to benchmark USV visual data privacy development in order to address the lack of techniques and datasets on this regard.

\subsection{Lack of foundation models for USVs} 
Transfer learning is the most common methodology for instructing deep learning models in the context of USV vision tasks. To elaborate, the deep learning network architectures were built on backbones that had previously been pre-trained on various datasets such as ImageNet and COCO. These datasets have broad applicability and may not accurately capture the characteristics of the maritime environment. Consequently, foundational models trained on large and diverse maritime datasets are critical for improving transfer learning in the development of USV models.

\subsection{Water texture} 
The water surface is, unsurprisingly, the most visible feature in the marine environment. The randomness of sea foam and waves makes accurate detection difficult. Furthermore, solar reflections cause random artifacts in conventional, thermal, and multispectral imaging technologies. The water surface appears to be uniform, but it varies greatly between bodies of water. Plants, animals, trash, and other complicating factors make detection more difficult. So, one of the future directions for USVs should be to address the stability of deep learning models of USVs that move in response to water surface fluctuations.

\subsection{Heterogeneity covered}
Heterogeneity is a significant challenge for deep learning development in USVs, as differences in maritime environmental conditions, object types, and object behaviors across nations make generalization difficult. Diverse water surfaces, cultural differences in maritime operation patterns, weather variations, and differences in visual features of objects (such as vessels, buoys, etc.) across water surface areas and countries are all significant sources of heterogeneity. This diversity necessitates extensive data collection and model adaptation to ensure that deep learning models perform well in a variety of settings without experiencing a significant drop in performance when exposed to unfamiliar conditions. Addressing heterogeneity frequently involves methods such as domain adaptation, transfer learning, and the integration of multi-domain datasets, but achieving consistent performance across heterogeneous environments remains a challenging task.

\section{Conclusion} \label{sec:conclude}
In this paper, we provide a comprehensive and systematic review of datasets and deep learning techniques for computer vision in USVs. We analyze the chronological synopsis as well as dataset statistics categorized by nation and region. We also compare these datasets with those from both visual tasks and sensor modalities employed in AVs due to the parallels between these two domains. Subsequently, we investigate USVs vision datasets for object identification and segmentation, along with other tasks, analyzing the most prevalent attributes such as quantity of samples, sensor modality, image resolution, FPS, quantity of annotated objects, open accessibility, FOV, area, and geographical location. Furthermore, we analyze the metadata and inventory of annotated objects from USVs vision datasets to elaborate on the ensuing challenges. We also engage in a wide range of deep learning techniques developed in USVs in the last 10 years, including object detection, segmentation, and multi-modal sensor fusion. Based on the analysis of both vision datasets and deep learning methodologies, we elaborate several insights and discussions, as well as highlight a number of challenges and prospective future directions in USVs vision.

\section{Acknowledgments} \label{section:ack}
The work was carried out in the framework of project INNO2MARE - Strengthening the Capacity for Excellence of Slovenian and Croatian Innovation Ecosystems to Support the Digital and Green Transitions of Maritime Regions  (Funded by the European Union under the Horizon Europe Grant N°101087348).


\printcredits

\bibliographystyle{cas-model2-names}

\bibliography{
    reference,
    bib_dirs/od,
    bib_dirs/seg/sem_seg,
    bib_dirs/seg/pap_seg,
    bib_dirs/seg/inst_seg,
    bib_dirs/fusion,
    bib_dirs/tracking,
    bib_dirs/other,
    bib_dirs/depth
}


\end{document}